\documentclass{article} 
\usepackage{collas2024_conference,times}
\usepackage{easyReview}
\usepackage{floatrow}
\usepackage{amsmath}
\usepackage{algorithm}
\usepackage{multirow}

\usepackage{booktabs}

\usepackage{xcolor,pifont}
\newcommand*\colourcheck[1]{%
  \expandafter\newcommand\csname #1check\endcsname{\textcolor{#1}{\ding{52}}}%
}

\usepackage{hyperref}
\hypersetup{
    colorlinks=true,
    linkcolor=blue,
    filecolor=magenta,      
    urlcolor=cyan,
}

\usepackage{algorithmicx} 
\usepackage{amsmath}
\usepackage{graphicx}
\usepackage{subcaption}
\usepackage{algpseudocode} 
\usepackage[normalem]{ulem}
\usepackage{hhline}
\usepackage{colortbl}

\useunder{ine}{}{}
\usepackage{wrapfig}
\usepackage{subfig}

\newfloatcommand{capbtabbox}{table}[][\FBwidth]

\usepackage{blindtext}

\usepackage{amsmath,amsfonts,bm}

\def\eqref#1{equation~\ref{#1}}

\def\1{\bm{1}}

\DeclareMathAlphabet{\mathsfit}{\encodingdefault}{\sfdefault}{m}{sl}
\SetMathAlphabet{\mathsfit}{bold}{\encodingdefault}{\sfdefault}{bx}{n}

\DeclareMathOperator*{\argmax}{arg\,max}

\usepackage{hyperref}
\hypersetup{
    colorlinks=true,
    linkcolor=red,
    filecolor=magenta,
    urlcolor=blue,
    citecolor=purple,
    pdftitle={Overleaf Example},
    pdfpagemode=FullScreen,
    }

\title{Towards Realistic Incremental Scenario in Class Incremental Semantic Segmentation}

\author{Jihwan Kwak$^{1}$, Sungmin Cha$^{2}$\;\& Taesup Moon$^{1,3}$\thanks{Corresponding author} \\
$^1$Department of Electrical and Computer Engineering, Seoul National University\\
$^2$Computer Science Department at the Courant Institute of Mathematical Sciences, New York University\\
$^3$ASRI / INMC / IPAI / AIIS, Seoul National University\\
\texttt{kkwakzi@snu.ac.kr}, \texttt{sungmin.cha@nyu.edu}, \texttt{tsmoon@snu.ac.kr} \\
}

\collasfinalcopy 

\begin{document}

\maketitle


\begin{abstract}

This paper addresses the unrealistic aspect of the \textit{overlapped} scenario, a commonly adopted incremental learning scenario in Class Incremental Semantic Segmentation (CISS). We highlight that the \textit{overlapped} scenario allows the \textbf{same} image to reappear in future tasks with different pixel labels, creating unwanted advantage or disadvantage to widely used techniques in CISS, such as pseudo-labeling and data replay from the exemplar memory. Our experiments show that methods trained and evaluated under the \textit{overlapped} scenario can produce biased results, potentially affecting algorithm adoption in practical applications. To mitigate this, we propose a practical scenario called \textbf{\textit{partitioned}}, where the dataset is first divided into distinct subsets representing each class, and then these subsets are assigned to corresponding tasks. This efficiently addresses the data reappearance artifact while meeting other requirements of CISS scenario, such as capturing the background shifts. Additionally, we identify and resolve the code implementation issues related to replaying data from the exemplar memory, previously overlooked in other works. Lastly, we introduce a simple yet competitive memory-based baseline, MiB-AugM, that handles background shifts in the exemplar memory. This baseline achieves state-of-the-art results across multiple tasks involving learning many new classes. Codes are available at \url{https://github.com/jihwankwak/CISS_partitioned}.
\end{abstract}

\section{Introduction}


Due to increasing industrial demands, recent studies have placed significant emphasis on understanding the behavior of models when learning from non-stationary streams of data. One area of particular interest is the Class Incremental Learning (CIL) problem, where a model learns new classes from incrementally arriving training data. The primary challenge in CIL lies in addressing the plasticity-stability dilemma~\citep{(spdilemma)carpenter87,(spdilemma)mermillod13}, whereby models must learn new concepts while mitigating the risk of \textit{catastrophic forgetting}~\citep{mccloskey1989catastrophic}, a phenomenon of inadvertently forgetting previously acquired knowledge when learning new concepts. To date, research in this field has expanded beyond methodological approaches \citep{kirkpatrick2017overcoming, lopez2017gradient, yoon2017lifelong} to include discussions on practical scenarios \citep{wu2019large, tao2020few, he2020incremental} that closely mirror real-world learning processes.


Motivated by its applications to autonomous driving and robotics, CIL has extended its reach to semantic segmentation tasks, known as Class Incremental Semantic Segmentation (CISS). In CISS, the model additionally encounters the challenge of  \textit{background shift} \citep{cermelli2020modeling}, which refers to a semantic drift of the background class between tasks. Specifically, since all pixels whose ground truth class does not correspond to the current task classes are annotated as background, objects from previous and future classes may be mislabeled as background. This exacerbates the forgetting of previous classes and hinders the knowledge acquisition of new classes.  

To apply and evaluate CISS methods, two incremental scenarios, \textit{disjoint} and \textit{overlapped} were initially introduced by \citet{cermelli2020modeling}. Since the \textit{disjoint} scenario failed to capture the background shift of unseen classes, most studies \citep{douillard2021plop, baek2022decomposed, zhang2022representation, zhang2023coinseg} have focused on the \textit{overlapped} scenario. Notably, recent methods have demonstrated state-of-the-art performance by either freezing parameters \citep{cha2021ssul, zhang2022mining} or applying strong regularization \citep{baek2022decomposed, zhang2023coinseg}. However, despite the active discussion on methodologies, there is a lack of awareness about the limitations of the \textit{overlapped} scenario, which have been overlooked until now.




\begin{figure}[t]
\begin{center}
\includegraphics[scale=0.4]{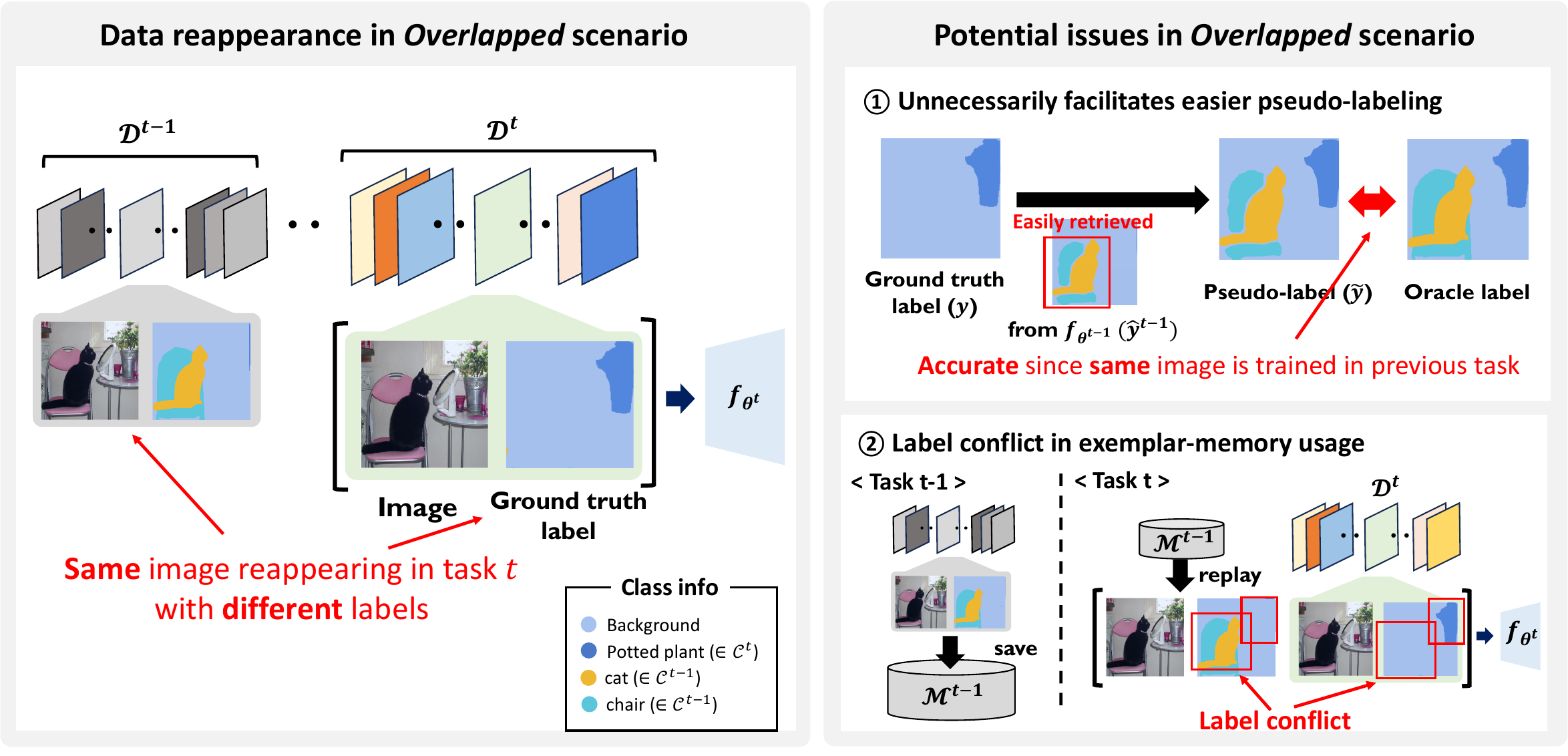}
\end{center}
\caption{(\textbf{Left}) Illustration of data reappearance issue in the \textit{overlapped} scenario. At task $t$, an image seen at task $t-1$ is given with different labels, which is far from a practical incremental learning scenario. (\textbf{Right}) Illustration of potential problems caused by the reappearance issue in the \textit{overlapped} scenario.}
\label{fig:motivation}

\end{figure}

We focus on addressing the unrealistic aspect of the widely adopted \textit{overlapped} scenario. As shown in Figure \ref{fig:motivation} (left), in the \textit{overlapped}, the \textbf{same} image can reappear in future tasks with \textbf{different} pixel labels. For example, an image seen in task $t-1$ labeled with \texttt{cat} and \texttt{chair} class can reappear at task $t$ labeled only with \texttt{potted plant} class. We term this as \textit{overlapping} data and show that this artifact can lead to unwanted advantage or disadvantage for certain techniques, resulting in biased outcomes for certain algorithms. Firstly, pseudo-labeling classes from the previous task becomes relatively easy for \textit{overlapping} data since the model has already seen the image from previous tasks. Secondly, if the \textit{overlapping} data saved in memory during the previous tasks are replayed with its saved labels, an image may be trained with two distinct labels, causing label conflicts. For example, in task $t$, the model learns the potted plant object with two labels: \texttt{potted plant} class from current data and \texttt{background} class from data replayed from the exemplar memory. Therefore, in the \textit{overlapped}, models unfairly benefit from learning pseudo-labels that are close to the oracle labels and encounter unnecessary label conflict when exemplar memory is utilized.



 
As an alternative, we propose a practical scenario, dubbed as \textit{partitioned}, that facilitates accurate and objective evaluation of CISS algorithms. This involves partitioning the dataset into mutually distinct subsets, each representing a specific class, and assigning each subset to its corresponding task. This methodology eliminates unnecessary artifact of the \textit{overlapped} scenario while satisfying the preconditions of the CISS scenario, such as capturing the background shifts of both previous and unseen classes. Furthermore, we address the overlooked code implementation issues in the exemplar memory, which have been ignored in prior studies. Prior code frameworks rather provide labels of the current task to the data replayed from the memory.

Lastly, motivated by the issues above, we introduce an efficient memory-based baseline, named MiB-AugM, that effectively handles the background shift of the current task class in the exemplar memory. Experiments with reproduced baselines show that our method outperforms state-of-the-art methods across several tasks involving the learning of numerous new classes at every task.

\section{Related works}








\subsection{Class incremental learning (CIL)}

\noindent{\textbf{Methodologies}} \ \ Research on CIL has primarily focused on image classification tasks, with methods falling into three main categories:  1) \textit{Regularization}-based \citep{kirkpatrick2017overcoming, chaudhry2018riemannian, ahn2019uncertainty, jung2020continual}, 2) \textit{Rehearsal}-based \citep{lopez2017gradient, shin2017continual, prabhu2020gdumb}, and 3) \textit{Architecture}-based \citep{rusu2016progressive, yoon2017lifelong, hung2019compacting, yan2021dynamically} solutions.
Among them, approaches that combine \textit{regularization} through knowledge distillation (KD) \citep{hinton2015distilling} and \textit{rehearsal} of exemplar memory \citep{rolnick2019experience} have achieved state-of-the-art performance \citep{li2017learning, buzzega2020dark}. Consequently, this integration has spurred a line of research dedicated to addressing by-product problems such as prediction bias \citep{wu2019large, ahn2021ss}.


\noindent{\textbf{Incremental scenarios}} \ \ Since the research demands of continual learning started from the non-stationary property of the real-world, several works have claimed the necessity of building up a practical incremental scenario that resembles the learning process of the real-world. For example, \citet{wu2019large, ahn2021ss} suggested an evaluation on \textit{large} scale CIL scenarios and \citet{hou2019learning, douillard2020podnet} proposed a scenario that includes a \textit{large base task} where the model has a chance to initially learn knowledge from many classes. 

With growing concerns about realistic scenarios that align with industrial demands, CIL studies have stretched out its application to settings with additional data constraints such as \textit{few-shot} \citep{tao2020few, yang2023neural} or \textit{online} \citep{he2020incremental, lin2023pcr}. Similarly, a new body of discussions on \textit{online}-CIL regarding practical scenarios \citep{koh2022online, chrysakis2023online} or constraints \citep{ghunaim2023real} has been introduced.

\subsection{Class incremental semantic segmentation (CISS)}\label{sec:related ciss}

\noindent{\textbf{Methodologies}}  \ \ Inspired by recent works of CIL, initial works in CISS \citep{cermelli2020modeling, douillard2021plop, michieli2021continual} took KD-based regularization as a general approach. MiB \citep{cermelli2020modeling} proposed a novel KD-based regularization to address the semantic drift of background label. PLOP \citep{douillard2021plop} brought the idea of \citet{douillard2020podnet} and utilized the feature-wise KD. On the other hand, few works \citep{yan2021framework, zhu2023continual} focused on employing the exemplar memory. \citet{yan2021framework} proposed to use a class-balanced memory but only focused on \textit{online} setting. \citet{zhu2023continual} proposed a memory sampling mechanism to ensure diversity among samples but has a reproduction issue\footnote{Code implementation not available}. Recently, SSUL \citep{cha2021ssul} which introduced a method to freeze the feature extractor have demonstrated outperforming performance. As a result, the recent state-of-the-art CISS methods have been focused on either freezing \citep{zhang2022mining} or regularizing the extractor with hard constraint \citep{zhang2023coinseg, chen2023saving}, 
while considering memory usage as an extra factor to enhance performance slightly.


\noindent{\textbf{Incremental scenarios}}  \ \
\citet{cermelli2020modeling} first established the learning scenarios of CISS, \textit{overlapped} and \textit{disjoint} on Pascal VOC 2012 \citep{everingham2010pascal} and ADE 20K \citep{zhou2017scene}. After many works \citep{douillard2021plop, baek2022decomposed} pointed out the impractical assumption of \textit{disjoint} where the existence of semantic shift of unseen class is excluded, current CISS research is actively being explored in \textit{overlapped} scenario. However, in contrast to CIL, where both practical scenarios and methodologies have been actively discussed, there has been no further discussion on the learning scenarios in CISS until now.

Building upon insights acquired from previous studies of CIL in classification, regularization-based methods have achieved impressive performance in CISS. However, previous studies have often overlooked the practical considerations in the learning scenarios. In this work, we point out an overlooked issue in the learning scenario (\textit{e.g.,} \textit{overlapped}) and propose a new scenario and a competitive baseline method that effectively leverages the exemplar memory.




\section{Problems of overlapped scenario and a proposed realistic scenario}

\subsection{Notation and problem setting}\label{sec:notation}

Class Incremental Learning (CIL) for image classification operates under the assumption that pairs of input data and its corresponding label for \textit{new} classes are accessible for training a model in each incremental \textit{task}.
At each task $t$, the model is trained on a new dataset, $\mathcal{D}^{t}$ annotated with a set of new classes $\mathcal{C}^{t}$. In the evaluation phase, the model is expected to distinguish between all the seen classes up to task $t$, denoted as $\mathcal{C}^{0:t}=\mathcal{C}^{0}\cup\dots\cup\mathcal{C}^{t}$. Each task is organized with disjoint classes, meaning there is no overlap of classes between the tasks, denoted as $\mathcal{C}^{i}\cap\mathcal{C}^{j} = \emptyset \text{ for } \forall i,j$.
Note that the model initially acquires knowledge of a large number of classes at the base task (task $0$), and gradually learns the remaining classes in subsequent tasks (task $1$ to $T$).


Class Incremental Semantic Segmentation (CISS) considers incremental learning in semantic segmentation, involving the pixel-level prediction of labels.
At task $t$, an input image of size $N$\footnote{For notational convenience, the image, originally considered as a $2\text{-}D$ array with height $(H)$ and width $(W)$, is now treated as a $1\text{-}D$ array with size $N$, where $N = H * W.$}, $x\in \mathbb{R}^{N \times 3}$, is paired with its corresponding ground-truth pixel labels $y\in \mathbb{R}^{N}$, denoted as $(x,y)\sim \mathcal{D}^{t}$. There exists at least one pixel that is annotated as one of the classes in $\mathcal{C}^{t}$ and pixels not belonging to $\mathcal{C}^{t}$ are labeled as the background label $c_{bg}$. Therefore, each task dataset $\mathcal{D}^{t}$ in CISS contains ground truth labels corresponding to $\mathcal{C}^{t}\cup{c_{bg}}$.
Note that objects from the \textit{past} or \textit{future} task's classes may be labeled as the background label in the current dataset $\mathcal{D}^{t}$, even though these same objects have their actual labels in other tasks. This label shift between tasks, known as \textit{background shift}, presents a significant challenge in CISS.
Previous works have considered two scenarios of CISS, \textit{disjoint} and \textit{overlapped}, which will be discussed in Section \ref{sec: CISS scenario intro}. Following previous works \citep{cha2021ssul, baek2022decomposed}, we also introduce an exemplar memory $\mathcal{M}^{t-1}$, which stores a subset of data from $\mathcal{D}^{0:t-1}$ and is used for task $t$. The size of memory remains consistent across the tasks, denoted as $|\mathcal{M}^{t}|=M \text{ }\forall t$.

At task $t$, our model, parameterized by  $\theta^{t}$, consists of a feature extractor and a classifier,  denoted as $f_{\theta^{t}}(\cdot)$. Given an input $x$, 
the model produces a consolidated score of seen classes for each pixel $z_{i}^{t}=f_{\theta^{t}}(x)_{i}\in\mathbb{R}^{|\mathcal{C}^{0:t}\cup \{ c_{bg}\}|}$.
The class prediction for each pixel is obtained by selecting the class with the highest score:
\begin{equation}
    \hat{y}_{i}^{t}=\argmax_{c\in\mathcal{C}^{0:t}\cup\{c_{bg}\}}z_{i,c}^{t},\label{eq:test}
\end{equation}
where $z_{i,c}^{t}$ represents the output logit (before softmax) of the $i^{\text{th}}$ pixel to the $c^{\text{th}}$ class in $\mathcal{C}^{0:t}\cup\{c_{bg}\}$.
\subsection{Incremental scenarios in CISS: disjoint and overlapped}\label{sec: CISS scenario intro}

\begin{figure}[t]
\begin{center}
\includegraphics[scale=0.4]{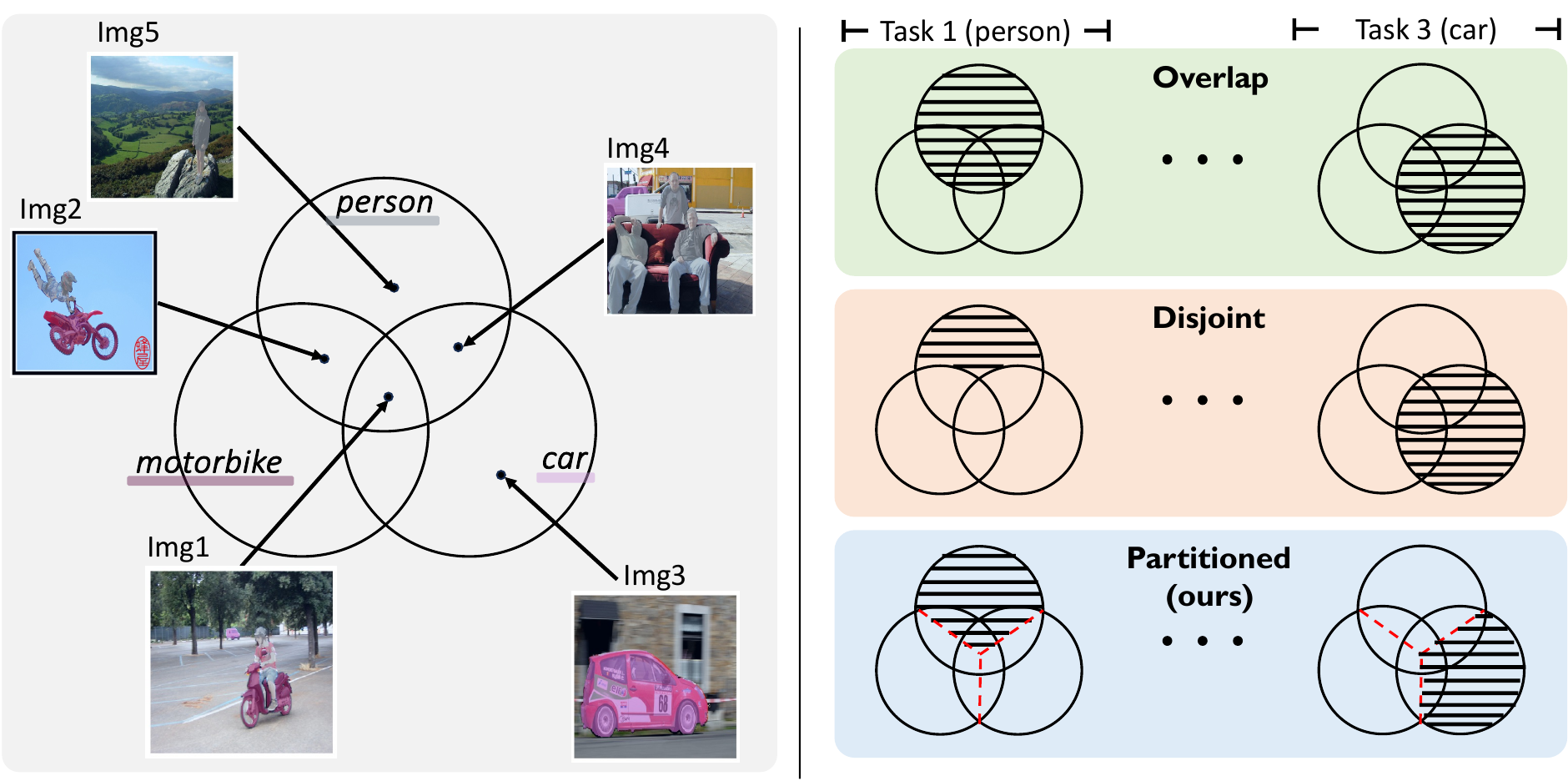}
\end{center}
\caption{(\textbf{Left}) A Venn diagram illustrating the relationship between ground truth class for each datum.
Each datum is assigned to a class set based on the object it contains. (\textbf{Right}) Comparison of $\mathcal{D}^{t}$ (highlighted with black line) for each scenario. Table \ref{tab: appendix dataset config details} in the Appendix provides a detailed summary of the assigned labels for each image in every incremental task.}
\label{fig:venn diagram}
\vspace{-5pt}
\end{figure}

Two incremental scenarios have been recognized as key scenarios of CISS~\citep{cermelli2020modeling}: \textit{disjoint} and \textit{overlapped}.
The dataset construction for each scenario can be compared using a set diagram, as depicted in Figure \ref{fig:venn diagram}.

Suppose we are constructing a CISS scenario using a dataset that has three object classes (\texttt{motorbike, car,}  and \texttt{person}), as annotated in Figure \add{\ref{fig:venn diagram}}. First, we define three sets representing each class, where each element in a set represents a datum (\textit{i.e.,} a pair of $(x,y)$).
Each datum is then assigned to one or more class sets based on its \textit{oracle} pixel labels. For example, the figure illustrates that \texttt{Img1} and \texttt{Img2} have objects whose pixels are labeled as \textit{\{car, motorbike, person\}} and \textit{\{motorbike, person\}}, respectively.
In this case, both \texttt{Img1} and \texttt{Img2} are elements of \texttt{motorbike} class set.
If an incremental scenario is defined to sequentially introduce the \texttt{person}, \texttt{motorbike}, and \texttt{car} classes as ground truth for each task, the Venn diagram of the three sets can be used to demonstrate the dataset given at each task in CISS. Namely, the data points covered by black lines indicate the dataset utilized for each task.



As illustrated in the right figure, \textit{disjoint} ensures the separation of the dataset for each task but necessitates prior knowledge of unseen classes to achieve this separation~\citep{cermelli2020modeling}. Moreover, \textit{disjoint} does not capture the background shift of classes from future tasks, leading to the widespread adoption of the \textit{overlapped} scenario in recent works~\citep{cermelli2020modeling,douillard2020podnet,cha2021ssul}. However, we argue that the \textit{overlapped} scenario has its own set of challenges.

\subsection{Unwanted advantage and disadvantage in the overlapped scenario}\label{sec:rethinking scenario}

The problem arises from the fact that, in \textit{overlapped}, previously seen images may be reintroduced in future tasks with different pixel labels, as illustrated in Figure \ref{fig:motivation}. For example, in task $t-1$, the pixels corresponding to the cat and chair object are labeled with their respective classes, while all other pixels are annotated as the \texttt{background} class. However, in task $t$, the \textbf{same} image may reappear with being labeled only with the \texttt{potted plant} class, leaving the remaining pixels, including the cat and chair object, as the \texttt{background} class. This phenomenon in \textit{overlapped} is far from a practical incremental learning scenario, and we term this problematic data as \textit{overlapping} data. In this section, we will show that overlapping data can create artificial advantage or disadvantage for techniques commonly used in current CISS, potentially resulting in misleading conclusions and impacting the development and adoption of algorithms in practical applications.

\noindent\textbf{Unwanted advantage of pseudo-labeling with previously learned model} \ \
To mitigate the issues arising from background shift caused by classes from the previous task, many recent studies \citep{douillard2021plop, cha2021ssul, zhang2023coinseg, chen2023saving} employ pseudo-labeling for the background region using predictions from the previously learned model. Formally, the pseudo-label $\tilde{y}_{i}$ can be defined as follows:\footnote{Since each study follows different rules for pseudo-labeling, we present the simplest format. Note that the core idea of pseudo-labeling is to use the predicted labels with a high prediction score, which is steered by hyper-parameter $\tau$.}
\begin{equation}
    \tilde{y}_{i}=\begin{cases}
    y_{i}  & \quad \text{if } y_{i} \in \mathcal{C}^{t}
    \\
    \hat{y}^{t-1}_{i}  & \quad    \text{if }  (y_{i}=c_{bg}) \land (s_{i}^{t-1} > \tau)
    \\
    c_{bg}  & \quad \text{else,}
    \end{cases}
    \label{ep:pseudo}
\end{equation}
where $s_{i}^{t-1}=\max_{c\in\mathcal{C}^{0:t-1}\cup \{c_{bg}\}}\frac{z_{i,c}^{t-1}}{\sum_{k}z_{i,k}^{t-1}}$ represents the output probability (after softmax) for $\hat{y}^{t-1}_{i}\in \mathcal{C}^{0:t-1}\cup\{c_{bg}\}$ and $\tau$ indicates the threshold for pseudo-labeling, respectively. An issue arises when the pseudo-labeling with the previously learned model is applied to overlapping data which is a pair of previously \textbf{seen} image and pixel labels annotated based on current task classes and the background class. The old class objects in the image were learned in previous tasks with the \textbf{same} image and corresponding label. Therefore, when pseudo-labeling this part with the past model, it produces accurate pseudo-labels, as shown in Figure \ref{fig:motivation} (right). This implies that overlapping data unnecessarily facilitates easier pseudo-labeling of classes from past task.

 


To showcase the aforementioned phenomenon,  we conducted experiments with the Pseudo-labeling Retrieval Rate (PRR) metric, defined as follows:
\begin{equation}
    \label{eq:pseudo metric}
    PRR (\hat{\mathcal{D}}, f_{\theta^{t-1}})=\frac{1}{|\hat{\mathcal{D}}|}\sum_{(x,y_{oracle})\sim \hat{D}}\text{mIoU}_{\mathcal{C}^{0:t-1}\cup\{c_{bg}\}}(y_{{oracle}}, \tilde{y}^{t-1})
\end{equation}

\begin{wraptable}{r}{0pt}
    \begin{minipage}{0.3\textwidth}
        \vspace{-20pt}
        \begin{table}[H]
        \resizebox{\textwidth}{!}{\begin{tabular}{c|cc}
        \toprule
        \textbf{} & \multicolumn{2}{c}{\textbf{\textit{PRR}} $(\cdot, f_{\theta^{0}}$)} \\
        Task      & $\mathcal{D}_{seen}$                  & $\mathcal{D}_{unseen}$                                \\ \midrule \midrule
        15-1      & 86.81                 & \textbf{66.91 (\textbf{19.90} $\downarrow)$}                \\
        15-5      & 84.15                 & \textbf{60.27 (\textbf{23.88} $\downarrow$)}
                     \\
        10-1      & 86.82                & \textbf{66.66 (\textbf{20.16} $\downarrow)$}                \\
        10-5      & 87.36                 & \textbf{68.40 (\textbf{18.96} $\downarrow)$}   
        \tabularnewline \bottomrule               
        \end{tabular}}
        \end{table}
        \vspace{-20pt}
        \caption{Pseudo-labeling Retrieval Rate (PRR) on $\mathcal{D}_{seen}$ and $\mathcal{D}_{unseen}$. The figure with the downarrow($\downarrow$) inside the parentheses indicates the difference between the two.}
        \label{tab:pseudo-labeling}
    \end{minipage}

\end{wraptable}



where $\hat{\mathcal{D}}$ indicates the dataset under evaluation that includes image $x$ with $y_{oracle}$ which indicates an oracle pixel labels containing all foreground and background classes. Also, $\tilde{y}^{t-1}$ denotes the pseudo-label of $x$ generated by $f_{\theta^{t-1}}$. To assess the retrieval rate of previous classes, Intersection-over-Unions (IoU) between $\tilde{y}^{t-1}$ and $y_{oracle}$ is averaged among previous classes, denoted as $\text{mIoU}_{\mathcal{C}^{0:t-1}\cup\{c_{bg}\}}(\cdot, \cdot)$.  For an in-depth description of the PRR metric, please refer to Figure \ref{fig:appendix PRR overview details} in the Appendix.

For verification, we modify the overlapped scenario by randomly dividing the overlapping data of two consecutive tasks, $\mathcal{D}^{0}\cap\mathcal{D}^{1}$, into two parts of the same size.  One part will be seen at the previous task, task $0$, denoted by $\mathcal{D}_{seen}$, and the other will not be seen, denoted by $\mathcal{D}_{unseen}$. As a result, at task $0$, the model is trained with $\mathcal{D}^{0} \setminus \mathcal{D}^{unseen}$. After training the model $f_{\theta^{0}}$, we evaluate the PRR results of $\mathcal{D}_{seen}$ and $\mathcal{D}_{unseen}$ by comparing IoU between oracle labels and pseudo-labels generated from $f_{\theta^{0}}$.

Table \ref{tab:pseudo-labeling} demonstrates PRR results across two different incremental tasks. Task $15\text{-}1$ indicates that the model initially learns $15$ classes and incrementally learns $1$ class at every task. Since the evaluation is done on task $1$, PRR is evaluated after learning $15$ classes at task $0$. For robustness, we report averaged results over $3$ random overlapping data splits, $\mathcal{D}_{seen}$ and $\mathcal{D}_{unseen}$ construction, and $3$ different class-ordering. For further details, please refer to Section \ref{sec: appendix pseudo analysis}

In Table \ref{tab:pseudo-labeling}, compared to PRR results of $\mathcal{D}_{seen}$, PRR results of $\mathcal{D}_{unseen}$ shows a notable decline in all tasks. For example, in $15\text{-}1$ task, PRR of $\mathcal{D}_{unseen}$ shows a $19.90$ decreased result compared to PRR of $\mathcal{D}_{seen}$. These experimental findings illustrate that the retrieval of previous labels is relatively straightforward when it is done on already seen overlapping data, which can cause unnecessary advantage to pseudo-labeling techniques.

\noindent\textbf{Unwanted disadvantage of using exemplar memory} Recent studies \citep{cha2021ssul, baek2022decomposed, zhu2023continual, chen2023saving} also actively utilize exemplar memory to store and replay past task data to alleviate forgetting. However, when overlapping data is saved in the memory, it can rather cause a \textit{label conflict} as illustrated in Figure \ref{fig:motivation} (right). For example, if the image with cat and chair objects labeled accordingly and the potted plant object marked as \texttt{background} is saved in task $t-1$ and replayed in task $t$, the model encounters the \textbf{same} image twice with disparate labels. For the potted plant object, data from the current task has the ground truth pixel label correctly labeled with \texttt{potted plant} class but the pixel labels from the memory are annotated as \texttt{background} class. This implies that overlapping data can worsen learning new concepts and remembering old knowledge when saved and replayed by the exemplar memory.

To observe the actual impact of the label conflict, we construct a following ablation experiment on the overlapping data in the exemplar memory. After training is done on the base task, class-balanced memory of size $M$ is constructed with $\mathcal{D}^{0}$ and is used to fine-tune the model along with current task data $\mathcal{D}^{1}$. Since overlapping data between $\mathcal{D}^{0}$ and $\mathcal{D}^{1}$ may exist in the exemplar memory, we denote this memory as \textit{overlapping memory} and report its average ratio. For comparison, we also construct \textit{non-overlapping memory} where overlapping data in the \textit{overlapping memory} is replaced with non-overlapping data in $D^{0}$. The model is expected to encounter label conflict when fine-tuning is done on $\mathcal{D}^{1}$ and \textit{overlapping memory} in contrast to the use of \textit{non-overlapping memory}. After training is done on the above two settings respectively, we compare the test mIoU to compare the impact of label conflict. For further implementation details, please refer to Section \ref{sec: appendix memory analysis}.

\begin{wraptable}{r}{0pt}
    \begin{minipage}{0.6\textwidth}
        \vspace{-20pt}
        \begin{table}[H]
      \resizebox{\textwidth}{!}{\begin{tabular}{cc|ccc}
      \toprule
\multicolumn{2}{c|}{\cellcolor[HTML]{FFFFFF}\textit{\textbf{Incremental Task}}} & \cellcolor[HTML]{FFFFFF}                                                                                                           & \multicolumn{2}{c}{\cellcolor[HTML]{FFFFFF}\textit{\textbf{Test overall mIoU}}}                                                                       \\ \cmidrule{1-2} \cmidrule{4-5} 
Task                                         & Class order                      & \multirow{-2}{*}{\cellcolor[HTML]{FFFFFF}\textit{\textbf{\begin{tabular}[c]{@{}c@{}}Overlapping ratio \\ in memory\end{tabular}}}} & \begin{tabular}[c]{@{}c@{}}Overlapping\\ memory\end{tabular} & \begin{tabular}[c]{@{}c@{}}Non-overlapping\\ memory\end{tabular} \\ \midrule \midrule
                                             & type a                           & 3\%                                                                                                                                & 50.02                                                                              & 50.15 (\textbf{0.13} $\uparrow$)                                            \\ 
\multirow{-2}{*}{15-1}                       & type b                           & 17\%                                                                                                                               & 24.26                                                                              & 27.02 (\textbf{2.56} $\uparrow$)                                            \\ \midrule
                                             & type a                           & 9\%                                                                                                                                & 50.30                                                                              & 51.85 (\textbf{1.55} $\uparrow$)                                            \\
\multirow{-2}{*}{15-5}                       & type b                           & 35\%                                                                                                                               & 35.13                                                                              & 37.41 (\textbf{2.28} $\uparrow$)                                            \\ \midrule
                                             & type a                           & 1\%                                                                                                                                & 71.98                                                                              & 72.27 (\textbf{0.29} $\uparrow$)                                            \\
\multirow{-2}{*}{10-1}                       & type b                           & 36\%                                                                                                                               & 25.19                                                                              & 27.35 (\textbf{2.16} $\uparrow$)                                            \\ \midrule
                                             & type a                           & 16\%                                                                                                                               & 43.94                                                                              & 45.38 (\textbf{1.44 $\uparrow$})                                            \\
\multirow{-2}{*}{10-5}                       & type b                           & 46\%                                                                                                                               & 36.21                                                                              & 39.31 (\textbf{3.10} $\uparrow$)                                     
\tabularnewline \bottomrule
\end{tabular}}
        \end{table}
        \vspace{-20pt}
        \caption{Test mIoU of models fine-tuned on $\mathcal{D}^{1}$ with each memory. Other than exemplar memory, we fix all the other training implementations. To reduce the randomness caused by memory construction, we report averaged results of models trained under $3$ different class-balanced memory.}
        \label{tab:memory analysis}
        \vspace{-20pt}
    \end{minipage}
    \vspace{-20pt}

\end{wraptable}

Table \ref{tab:memory analysis} shows the test mIoU of models fine-tuned on $\mathcal{D}^{1}$ with each memory. Test mIoU of the model increases in all incremental tasks when \textit{overlapping memory} is replaced with \textit{non-overlapping memory}. This increase is further amplified when overlapping ratio gets higher, which implies that the overlapping data in the memory results in label conflict in the \textit{overlapped} scenario.

Through the above experiments, we have shown that the unrealistic artifact of the \textit{overlapped} scenario, namely overlapping data, is not negligible since it may lead to biased results toward several techniques. Therefore, in the following section, we will introduce a new scenario that handles the overlapping issue while satisfying other conditions of CISS scenario.

\subsection{Proposed scenario: Partitioned} \label{sec:proposed}



The issue of \textit{overlapped} primarily arises from overlapping data that provide different labels across various tasks. Therefore, we reconsider the disjointness property in CISS scenario by suggesting a scenario that meets the core requirement of \textit{disjoint} and \textit{overlapped}: 1) capturing background shifts of both previous and unseen classes 2) disjointness for eliminating overlapping data. our proposed scenario can be summarized in two steps: 1) partitioning the dataset into distinct subsets representing each class and 2) assigning each class subset to a corresponding task dataset.



\begin{wraptable}{r}{0pt}
    \vspace{-20pt}
    \begin{minipage}{0.45\textwidth}
        \vspace{-20pt}
        \begin{table}[H]
        \resizebox{\textwidth}{!}{\begin{tabular}{c|cc}
        \toprule
    \begin{tabular}[c]{@{}c@{}}Incremental \\ scenario \end{tabular}       & \begin{tabular}[c]{@{}c@{}}Removal of \\ overlapping data \end{tabular} & \begin{tabular}[c]{@{}c@{}} Capturing \\ background shifts   \end{tabular} \\ \midrule \midrule
\textit{disjoint}           &    \textcolor{green}{\ding{51}}                                                                    &  \textcolor{red}{\ding{55}}                                                                        \\ 
\textit{overlapped}         &        \textcolor{red}{\ding{55}}                                                                &   \textcolor{green}{\ding{51}}                                                                       \\ 
\textit{partitioned} (ours) &     \textcolor{green}{\ding{51}}                                                                   &   \textcolor{green}{\ding{51}}  
\tabularnewline \bottomrule
\end{tabular}}
            
            \caption{Comparison of scenarios including \textit{partitioned} in two criteria towards realistic CISS scenario.}
            \label{tab: partitioned advantages}
        \end{table}
    \vspace{-20pt}    
    \end{minipage}

\end{wraptable}

\begin{figure}[t]
\begin{center}
\includegraphics[scale=0.35]{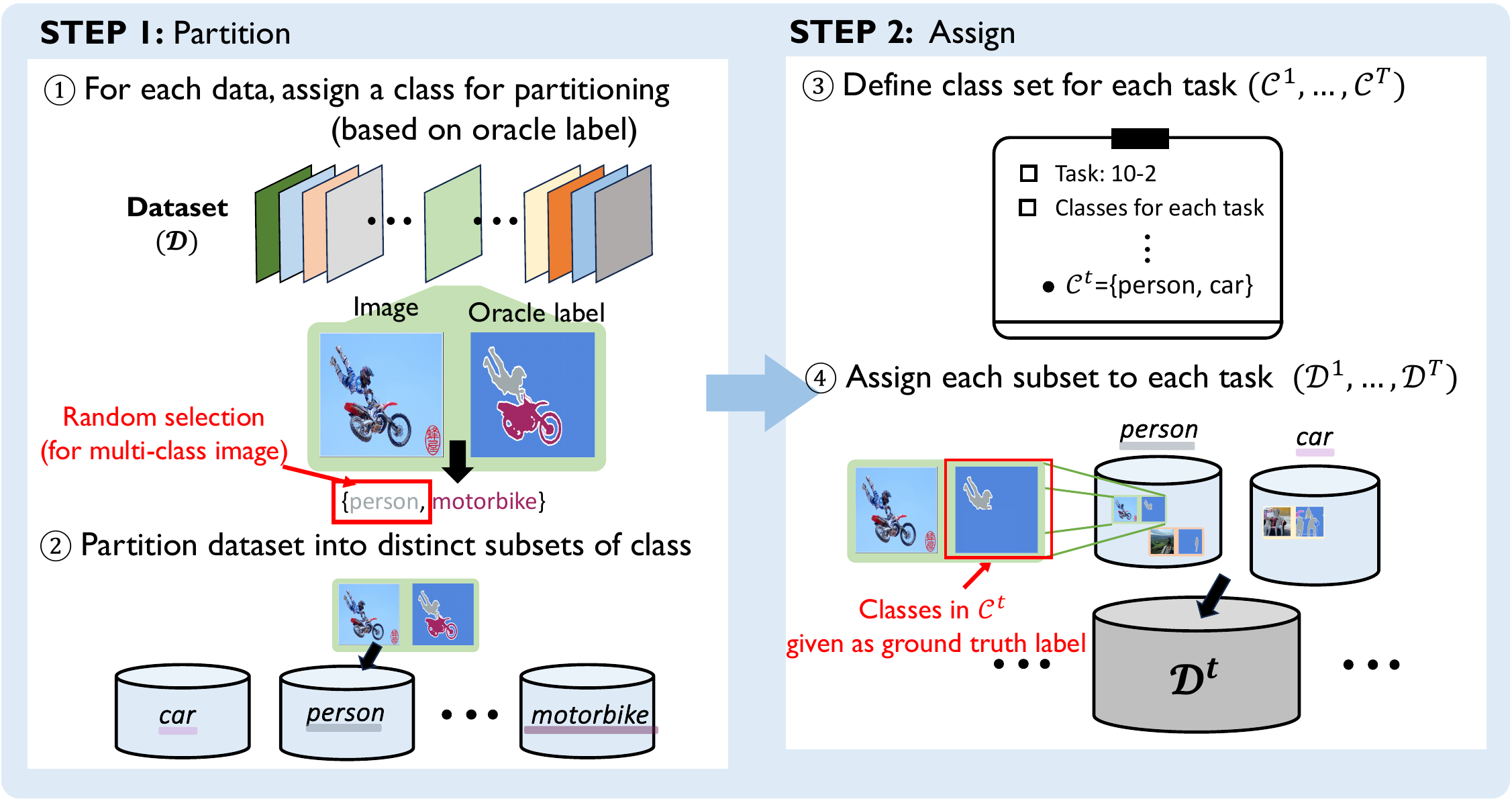}
\end{center}
\caption{Illustration of dataset construction steps for \textit{partitioned} scenario.
}
\label{fig:partitioned}
\vspace{-5pt}
\end{figure}

Figure \ref{fig:partitioned} demonstrates the process of constructing the proposed \textit{\textbf{partitioned}} scenario. First, for each data (image-pixel labels pair), we assign a class for partitioning based on the oracle pixel labels. If the oracle pixel labels consist of at least two classes, we randomly select one class. Experimental results in the later section show that baseline results are robust to this random selection. Second, we partition the whole dataset ($\mathcal{D}$) into distinct subsets based on the assigned class for partitioning. Third, following the required task information (\textit{e.g.,} task $10\text{-}2$), we define the newly learned class set for each task ($\mathcal{C}^{0}, ..., \mathcal{C}^{T}$). Finally, we assign each subset to the corresponding task dataset ($\mathcal{D}^{0}, ..., \mathcal{D}^{T}$).

As summarized in Table \ref{tab: partitioned advantages}, our proposed \textit{partitioned} 1) guarantees disjointness between task datasets which eliminates overlapping data and 2) captures the background shift of both previous and unseen classes. Figure \ref{fig:venn diagram} (right) also visualizes the dataset for each task in the Venn diagram for comparing \textit{partitioned} with \textit{disjoint} and \textit{overlapped}. Experimental results on \textit{partitioned} with reproduced baselines are reported in Section \ref{sec:experiment}.

\subsection{New baseline for exemplar memory replay}\label{sec:method}


Since the issue of overlapping data is mitigated in our proposed scenario, 
it is natural to expect an effective usage of memory in \textit{partitioned}. Therefore, we propose a simple yet competitive baseline that integrates MiB \citep{cermelli2020modeling} with an extra loss function, tailored for the case of using exemplar memory. 

Following notations in MiB \citep{cermelli2020modeling}, the overall loss function of our method at task $t$ is defined as:
\begin{equation}
    \label{eq:method}
    \begin{aligned}
     \mathcal{L}(\theta^{t})  =\underbrace{\frac{1}{|\mathcal{D}^{t}|}\sum_{(x,y)\in \mathcal{D}^{t}}\mathcal{L}_{unce}(y, f_{\theta^{t}}(x)) +\frac{\lambda}{|\mathcal{D}^{t}\cup\mathcal{M}^{t-1}|}\sum_{(x,y)\in \mathcal{D}^{t}\cup\mathcal{M}^{t-1}}\mathcal{L}_{unkd}(f_{\theta^{t-1}}(x), f_{\theta^{t}}(x))}_\text{From MiB \citep{cermelli2020modeling}} \\  + \underbrace{\frac{1}{|\mathcal{M}^{t-1}|}\sum_{(x,y)\in\mathcal{M}^{t-1}}\mathcal{L}_{mem}(y, f_{\theta^{t}}(x))}_\text{Our proposed memory loss}
    \end{aligned}
\end{equation}
where $\mathcal{L}_{unce}(\cdot)$ and $\mathcal{L}_{unkd}(\cdot)$ are the loss functions employed in MiB\footnote{The original paper utilizes the terminologies $\mathcal{L}_{ce}(\cdot)$ and $\mathcal{L}_{kd}(\cdot)$ to represent loss functions. However, to avoid potential confusion with conventional cross-entropy and knowledge distillation losses, we choose to adjust the terminology.}. For the specific formulation of $\mathcal{L}_{unce}(\cdot)$ and $\mathcal{L}_{unkd}(\cdot)$ in our notation, please refer to Section \ref{sec: appendix MiB-M, DKD-M, PLOP-M}.
Regarding $\mathcal{L}_{mem}(\cdot)$, we suggest utilizing cross-entropy loss with the augmented predictions $\dot{p}_{i,c}^{t}$ as follows:
\begin{equation}
    \mathcal{L}_{mem}(y, f_{\theta^{t}}(x))=-\frac{1}{N}\sum_{i=1}^{N}\log \dot{p}^{t}_{i,y_{i}}.
    \label{eq:mem loss}
\end{equation}
Here, $\dot{p}^{t}_{i,y{i}}$ represents the augmented output probability (after softmax) for the ground truth label of a $i^{th}$ pixel. Moreover, the augmented prediction $\dot{p}_{i,c}^{t}$ can be defined as follows:
\begin{equation}
    \dot{p}_{i,c}^{t}=\begin{cases}
    p_{i,c}^{t}  & \quad \text{if } c \neq c_{bg}
    \\
    p_{i,c_{bg}}^{t}+\sum_{k\in\mathcal{C}^{t}}p_{i,k}^{t}
      & \quad    \text{if }  c=c_{bg},
    \end{cases}
    \label{eq:mem_prediction}
\end{equation}
where $p^{t}_{i,c}={\exp^{z^{t}_{i,c}}} / {\sum_{k\in\mathcal{C}^{0:t}\cup \{c_{bg}\}}\exp^{z^{t}_{i,k}}}$ is the output probability (after softmax) for class $c$ using $f_{\theta^{t}}$. 

Note that the motivation behind employing Equation (\ref{eq:mem loss}) stems from the background shift in $\mathcal{M}^{t-1}$. 
For example, consider data stored in the exemplar memory containing objects belonging to the class of the current task. Since this data only has classes from past tasks as a ground truth label, objects of the current task are annotated as \texttt{background}. This background shift of the new class confuses the acquisition of knowledge of classes from new tasks. To mitigate this conflict, we adopt the intuition of \citet{zhang2022mining}, wherein both background and unseen classes are treated as the same class. 
Additionally, inspired by the prediction augmentation technique discussed in \citet{cermelli2020modeling}, we aggregate the prediction values of $c_{bg}$ and $c \in \mathcal{C}^{t}$, specifically in the case of $c=c_{bg}$ in Equation (\ref{eq:mem_prediction}). 
Note that this allows for a positive update of prediction scores for both background and new classes when the model is trained with pixels labeled as \texttt{background}, thereby alleviating the label conflict of the data in $\mathcal{M}^{t-1}$. 



\subsection{Implementation issues in previous CISS studies}\label{sec:implementation}



In addition to the issues discussed in the previous sections, we will highlight a certain overlooked aspect of code implementation errors in CISS studies.
It is worth noting that the labeling implementation for data replayed from the exemplar memory is incorrect in recent studies \citep{cha2021ssul, baek2022decomposed, zhang2022mining, zhang2023coinseg}. Upon examination of their official code repositories, it becomes evident that the data from the exemplar memory \textbf{do not provide} ground truth labels for previous objects; instead, they only assign ground truth labels based on $\mathcal{C}^{t}$ or $c_{bg}$ (refer to Figure \ref{fig:appendix target labeling for memory error} in the Appendix). We would like to emphasize the importance of accurate implementation as such discrepancies in implementation can lead to inherent experimental biases. To address this, we rectified the error in the implementation and conducted all experiments with the corrected version, of which codes are at \ \url{https://github.com/jihwankwak/CISS_partitioned}.



We believe that addressing these implementation corrections and providing reproduced baselines constitutes a valuable contribution of our work, which can benefit future researchers in CISS studies.


 

\section{Experimental results}\label{sec:experiment}
\subsection{Experimental setups}


\noindent{\textbf{Dataset and protocols}}  \ \ We evaluated our methods based on two incremental scenarios, namely \textit{overlapped} and \textit{partitioned} (ours), using Pascal VOC 2012 \citep{everingham2010pascal} dataset. To assess the overall performance, we conducted evaluations across various tasks (\textit{e.g.,} $15\text{-}1$ task) with differing characteristics (\textit{e.g.,} large/small base tasks). Note that the total number of training data for all the tasks is different between \textit{overlapped} and \textit{partitioned} due to disjointness property of the \textit{partitioned}. Detailed data configurations for two scenarios are provided in Table \ref{tab:appendix training data numbers}.

\noindent{\textbf{Evaluation metrics}}  \ \ We utilize the mean Intersection-over-Union (mIoU) as our evaluation metric, which represents the averaged IoU over defined classes. The range of the average is provided in the table, such as $1\text{-}15$ and $16\text{-}20$, distinguishing between base classes and incrementally learned classes. Following \citet{cermelli2020modeling}, the mIoU of the \texttt{background} class is only included in the \textit{all} category, as it exists in both the base task and incremental tasks.

\noindent{\textbf{Baseline}}  \ \ Since DKD \citep{baek2022decomposed} stands as one of the state-of-the-art methods in CISS, we reproduced the general regularization approaches, MiB \citep{cermelli2020modeling} and  PLOP \citep{douillard2021plop}, based on its implementation. Note that pseudo-labeling is applied in PLOP and is not applied in DKD and MiB. The detailed implementation of memory usage, such as PLOP-M, $\text{DKD-M}^{\dagger}$ \footnote{Implementation of DKD with memory is modified from the original implementation, dubbed as $\text{DKD-M}^{\dagger}$ as the original implementaiton, DKD-M, cannot be directly used after memory modification mentioned in Section \ref{sec:implementation}}, is explained in Section \ref{sec: appendix MiB-M, DKD-M, PLOP-M}. In this work, we assume the non-usage of the off-the-shelf detectors, thus we do not compare with the latest methods such as MicroSeg \citep{zhang2022mining} and CoinSeg \citep{zhang2023coinseg}.

\noindent{\textbf{Implementation details}}  \ \ For all experiments, following \citet{cermelli2020modeling}, we employed a DeepLab v3 segmentation network \citep{chen2017rethinking} with a ResNet-101 \citep{he2016deep} backbone, pre-trained on ImageNet \citep{deng2009imagenet}. For training our method (MiB-AugM), we optimize the network with the learning rate of $10^{-3}$ for the backbone model and $10^{-2}$ for the rest. SGD with a Nesterov momentum value of $0.9$ is used for optimization in all incremental steps. Other training implementations are equal to details written in \citet{cermelli2020modeling}. For detailed implementations of our model and other baselines, please refer to Section \ref{sec: appendix hyperparameters}. In terms of exemplar memory, consistent with prior research \citep{cha2021ssul, baek2022decomposed}, we utilized memory with a fixed size of $M=100$.

\begin{table}[t]
\centering
\resizebox{\textwidth}{!}{
\begin{tabular}{l|cccccccccccc}
\toprule
\multirow{3}{*}{} & \multicolumn{12}{c}{\textbf{\textit{Overlapped}}}                                                                                                                                                                                 \\ \cmidrule{2-13} 
                  & \multicolumn{3}{c|}{15-1 Task}                                        & \multicolumn{3}{c|}{5-3 Task}                                         & \multicolumn{3}{c|}{10-1 Task}             & \multicolumn{3}{c}{10-5 Task}                    \\
                  & 1-15           & 16-20          & \multicolumn{1}{c|}{all}            & 1-5           & 6-20          & \multicolumn{1}{c|}{all}            & 1-10   & 11-20  & \multicolumn{1}{c|}{all}   & 1-10            & 11-20           & all            \\ \midrule \midrule
MiB  \scriptsize \citep{cermelli2020modeling}             & 31.92          & 17.33          & \multicolumn{1}{c|}{30.98}          & 60.72          & 49.62          & \multicolumn{1}{c|}{53.94}          & 9.15  & 22.42 & \multicolumn{1}{c|}{18.93} & 67.77          & 58.16          & 64.25          \\
PLOP \scriptsize \citep{douillard2021plop}             & 62.77          & 12.25          & \multicolumn{1}{c|}{49.99}          & 17.46          & 36.08          & \multicolumn{1}{c|}{34.05}          & 23.53 & 11.66 & \multicolumn{1}{c|}{17.26} & 58.78          & 50.39          & 56.29          \\
PLOP-M \scriptsize \citep{douillard2021plop}           & 63.71          & 25.85          & \multicolumn{1}{c|}{55.00}          & 65.02          & 43.41         & \multicolumn{1}{c|}{50.64}          & 15.97 & 9.78  & \multicolumn{1}{c|}{15.32} & 74.41          & 56.53          & 66.71          \\
DKD \scriptsize \citep{baek2022decomposed}              & 76.23          & 40.03          & \multicolumn{1}{c|}{68.14}          & 64.42          & 51.08          & \multicolumn{1}{c|}{56.01}          & 70.79 & 46.08 & \multicolumn{1}{c|}{59.67} & 71.33          & 59.56          & 66.64          \\
$\text{DKD-M}^{\dagger}$ \scriptsize \citep{baek2022decomposed}            & \textbf{76.66} & \textbf{45.29} & \multicolumn{1}{c|}{\textbf{69.84}} & \textbf{68.14} & 54.15 & \multicolumn{1}{c|}{59.14} & \textbf{72.50} & \textbf{53.14} & \multicolumn{1}{c|}{\textbf{64.04}} & 72.19 & 59.80 & 67.18 \\
MiB + AugM (ours)        & 73.21          & 36.20          & \multicolumn{1}{c|}{65.06}          & 67.93          & \textbf{59.42}          & \multicolumn{1}{c|}{\textbf{62.76}}          & 64.02 & 37.89 & \multicolumn{1}{c|}{52.30} & \textbf{74.53}          & \textbf{61.77}          & \textbf{69.19}          \\ \bottomrule \toprule
                  & \multicolumn{12}{c}{\textbf{\textit{Partitioned}}}                                                                                                                                                                                \\ \cmidrule{2-13} 
                  & \multicolumn{3}{c|}{15-1 Task}                                        & \multicolumn{3}{c|}{5-3 Task}                                         & \multicolumn{3}{c|}{10-1 Task}             & \multicolumn{3}{c}{10-5 Task}                    \\
                  & 1-15           & 16-20          & \multicolumn{1}{c|}{all}            & 1-5           & 6-20          & \multicolumn{1}{c|}{all}            & 1-10  & 11-20 & \multicolumn{1}{c|}{all}   & 1-10           & 11-20          & all            \\ \midrule \midrule
MiB \scriptsize \citep{cermelli2020modeling}              & 22.45          & 13.04          & \multicolumn{1}{c|}{23.01}          & 50.47          & 45.65          & \multicolumn{1}{c|}{48.71}          & 2.58  & 16.53 & \multicolumn{1}{c|}{12.88} & 62.42          & 54.61          & 60.00          \\
PLOP \scriptsize \citep{douillard2021plop}              & 63.62          & 11.72          & \multicolumn{1}{c|}{49.12}          & 16.48          & 27.66          & \multicolumn{1}{c|}{27.67}          & 14.85 & 9.90  & \multicolumn{1}{c|}{11.79} & 49.86          & 44.96          & 49.48          \\
PLOP-M \scriptsize \citep{douillard2021plop}            & 63.64          & 23.48          & \multicolumn{1}{c|}{54.45}          & 57.30          & 38.50          & \multicolumn{1}{c|}{45.22}          & 13.72 & 12.51 & \multicolumn{1}{c|}{15.55} & 70.04          & 51.75          & 62.36          \\
DKD \scriptsize \citep{baek2022decomposed}              & 73.86          & 35.86          & \multicolumn{1}{c|}{65.51}          & 61.72          & 47.92          & \multicolumn{1}{c|}{53.14}          & 67.24 & 43.11 & \multicolumn{1}{c|}{56.67} & 64.88          & 55.68          & 64.12          \\
$\text{DKD-M}^{\dagger}$ \scriptsize \citep{baek2022decomposed}             & \textbf{76.82} & \textbf{44.40} & \multicolumn{1}{c|}{\textbf{69.79}} & \textbf{66.69}          & 52.16          & \multicolumn{1}{c|}{57.40}          & \textbf{70.91} & \textbf{52.03} & \multicolumn{1}{c|}{\textbf{62.80}} & 70.65          & 56.53          & 64.92          \\
MiB + AugM (ours)       & 73.15          & 34.02          & \multicolumn{1}{c|}{64.29}          & 65.47          & \textbf{59.57}          & \multicolumn{1}{c|}{\textbf{62.34}}          & 63.85 & 41.24 & \multicolumn{1}{c|}{53.82} & \textbf{72.59}          & \textbf{59.36}          & \textbf{67.13}         
\tabularnewline \bottomrule
\end{tabular}}
\caption{Test mIoU results after the final incremental task, averaging the results over $3$ different runs.}
\label{tab:results}
\end{table}

\subsection{Main results}


\subsubsection{Key observations of partitioned scenario via baseline results} \label{sec: exp-memory overllok}

\noindent{\textbf{Label conflicts of the exemplar memory resolved in the \textit{partitioned} scenario}} \ \ Table \ref{tab:results} reports the mIoU results of test data after training the final incremental task. In every tasks, all methods demonstrate a greater improvement when exemplar memory is used in the \textit{partitioned} compared to the \textit{overlapped}. For example, in $5\text{-}3$ task, MiB, PLOP, and DKD shows $8.82$, $16.59$, and $3.13$ gains respectively in the \textit{overlapped} while $13.63$, $17.55$, and $4.26$ improvements were observed in the \textit{partitioned}, which is also visualized in Figure \ref{fig:memory gain (5-3)}. Moreover, in $10\text{-}1$ task, PLOP demonstrates a decline from $17.26$ to $15.32$, while an increase of $3.76$ is observed in the \textit{partitioned}. These results implies that, in the \textit{overlapped} scenario, label conflicts caused by the exemplar memory impacts the final results of current CISS methods as our previous observations, and that the \textit{partitioned} scenario can efficiently resolve this issue, enabling unbiased comparison between methods.

\noindent{\textbf{Robust to random selection in constructing the \textit{partitioned} scenario }}
Given that the data splitting rule used in \textit{partitioned} entails randomness, we analyze the impact of this variance in terms of the final results of each method trained on different dataset construction seeds. We also report the variance of models learned on \textit{overlapped} that does not involve randomness for data construction to represent the default variance for training each method. The figures displayed above the bar plot in Figure \ref{fig:variance (5-3} illustrate the performance variance of each method. Notably, the models trained on the \textit{partitioned} dataset exhibit similar randomness to the default variance observed in the \textit{overlapped} scenario. This suggests that the final results of each model are robust to the randomness during the construction of the \textit{partitioned} dataset.

\subsubsection{Experimental results on MiB-AugM}
\begin{figure}[t]
    \centering
    \resizebox{\textwidth}{!}{\begin{subfigure}{0.3\textwidth}
        \centering
        \includegraphics[width=\linewidth]{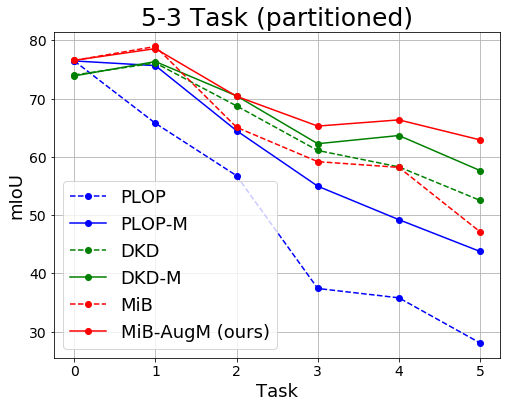}
            
        
        \caption{mIoU on each incremental tasks}
        \label{fig: each task (5-3)}
    \end{subfigure}
    \begin{subfigure}{0.3\textwidth}
        \centering
        \includegraphics[width=\linewidth]{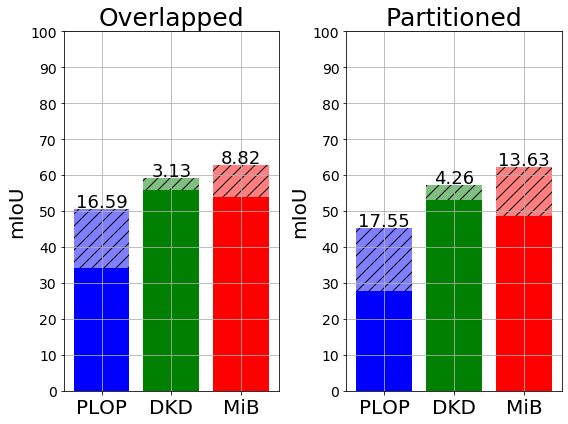}
        \vspace{0.005cm}
        \caption{Memory gain of each method}
        \label{fig:memory gain (5-3)}
    \end{subfigure}
    \begin{subfigure}{0.3\textwidth}
        \centering
        \includegraphics[width=\linewidth]{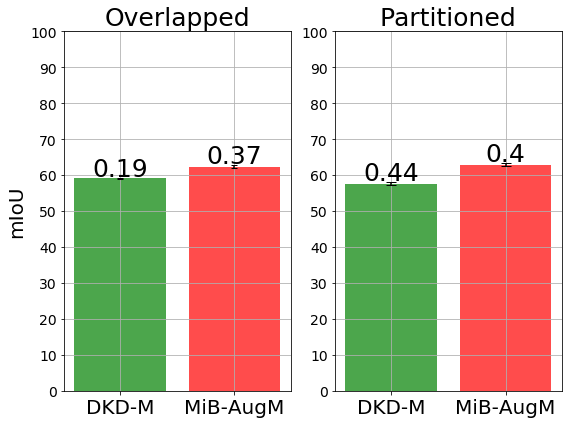}
        \vspace{0.005cm}
        \caption{Variance for each setting (3 seeds)}
        \label{fig:variance (5-3}
    \end{subfigure}}

    \caption{Results of each method on 5-3 Task}
    \label{fig: results different seeds}
\end{figure}

\noindent{\textbf{MiB-AugM as a new baseline for plasticity}} \ \ MiB with our proposed memory loss, MiB-AugM, exhibits state-of-the-art results across several incremental tasks.  In Table \ref{tab:results}, in both the \textit{partitioned} $5\text{-}3$ and $10\text{-}5$ tasks, MiB-AugM outperforms the state-of-the-art method by $4.94$ and $2.21$, respectively. Additionally, Figure \ref{fig: each task (5-3)} demonstrates that this superiority remains consistent for every task.

While our proposed memory-based baseline may not demonstrate superior results in every task, we highlight two key points that support MiB-AugM as a promising future baseline for plasticity. Firstly, MiB-AugM shows better performance in tasks that are more complex in terms of learning new concepts. As more classes are introduced in the new task, the model must distinguish not only between old and new classes but also among the new classes themselves. Hence, a higher level of plasticity is required, particularly in the $5\text{-}3$ and $10\text{-}5$ tasks. Secondly, MiB-AugM is an efficient method in terms of hyper-parameters. Unlike other methods that necessitate multiple hyper-parameters (at least $4$ for DKD and PLOP), MiB-AugM has only $\lambda$. Given that MiB-AugM can efficiently achieve state-of-the-art performance in several incremental tasks requiring complex plasticity, we insist that it holds promise as a valuable baseline for plasticity in future CISS studies.


\section{Concluding Remarks, Limitation, and Future work}

Our work addresses the unrealistic aspect of the \textit{overlapped} setting in CISS, where identical images are reused in future tasks with different pixel labels, which is not practical for real-world learning scenarios. We demonstrate that this artifact can lead to unfair advantage or disadvantage for commonly used techniques in CISS, resulting in biased comparisons among algorithms. To address this, we propose an alternative \textit{partitioned} scenario that eliminates data reappearnce while meeting incremental learning requirements, such as capturing background shifts between previous and new classes. Additionally, we introduce a simple yet competitive exemplar memory-based method that effectively handles background shifts in stored data. This method efficiently uses exemplar memory in the proposed setting and outperforms state-of-the-art methods on several tasks that involve learning multiple new classes incrementally.

\noindent{\textbf{Limitation and broader impact}} While promising, our work has a few limitations that warrant future exploration. Firstly, our proposed replay-based baseline exhibits inferior performance compared to state-of-the-art methods in the $15\text{-}1$ and $10\text{-}1$ tasks, which involve learning a small number of classes in common. Addressing the stability-plasticity dilemma regarding the number of new classes requires further research. Secondly, our findings are based on experiments conducted solely on the Pascal VOC dataset, which is relatively small. Future research should explore other datasets beyond Pascal VOC \citep{everingham2010pascal} or ADE 20K \citep{zhou2017scene}, with larger volumes of data. We intend to investigate incremental scenarios involving larger datasets as part of our future work.

\newpage

\section*{Acknowledgements}

This work was supported in part by the National Research Foundation of Korea (NRF) grant [No.2021R1A2C2007884] and by Institute of Information \& communications Technology Planning \& Evaluation (IITP) grants
[RS-2021-II211343, RS-2021-II212068, RS-2022-II220113,
RS-2022-II220959] funded by the Korean government (MSIT). It was also supported by SNU-NAVER Hyperscale AI Center and AOARD Grant No. FA2386-23-1-4079.

\bibliography{collas2024_conference}
\bibliographystyle{collas2024_conference}

\newpage
\appendix
\section{Appendix}

\subsection{Data configuration comparison: Overlap, Disjoint, and Partitioned}


The tables in Table \ref{tab: appendix dataset config details} provide a summary of the ground truth labels for each image in every incremental tasks. The first two columns show the image name and its corresponding oracle classes, while subsequent columns display the ground truth labels for each task. It's important to note that pixels not belonging to classes in $\mathcal{C}^{t}$ are annotated as the \texttt{background} class (abbreviated as \textit{bg} in the table). To illustrate background shifts, pixels affected by unseen classes are highlighted in red, while those affected by previous classes are highlighted in blue. A hatched line is used for data not present in each task.


\begin{table}[!htb]
    \caption{Labeling information of each image in Figure \ref{fig:venn diagram} for each incremental scenario: \textit{overlapped}, \textit{Disjoint}, and \textit{Ours}. }
    \label{tab: appendix dataset config details}
    \begin{subtable}{.33\linewidth}
      \centering
        \caption{Overlapped}
        \resizebox{0.95\textwidth}{!}{\begin{tabular}{cc|ccc}
\hline
\cellcolor[HTML]{EFEFEF}                                 & \cellcolor[HTML]{EFEFEF}                                        & \multicolumn{3}{c}{\cellcolor[HTML]{EFEFEF}\textbf{Given ground truth class}}                                                                                                     \\
\multirow{-2}{*}{\cellcolor[HTML]{EFEFEF}\textbf{Image}} & \multirow{-2}{*}{\cellcolor[HTML]{EFEFEF}\textbf{Oracle class}} & \begin{tabular}[c]{@{}c@{}}Task 1\\ (person)\end{tabular} & \begin{tabular}[c]{@{}c@{}}Task 2\\ (motorbike)\end{tabular} & \begin{tabular}[c]{@{}c@{}}Task 3\\ (car)\end{tabular} \\ \hline
                                                         & person                                                          & person                                                    & {\color[HTML]{3531FF} bg}                                    & {\color[HTML]{3531FF} bg}                              \\
                                                         & motorbike                                                       & {\color[HTML]{FE0000} bg}                                 & motorbike                                                    & {\color[HTML]{3531FF} bg}                              \\
\multirow{-3}{*}{Img1}                                   & car                                                             & {\color[HTML]{FE0000} bg}                                 & {\color[HTML]{FE0000} bg}                                    & car                                                    \\ \hline
                                                         & person                                                          & person                                                    & {\color[HTML]{3531FF} bg}                                    &                                                        \\
\multirow{-2}{*}{Img2}                                   & motorbike                                                       & {\color[HTML]{FE0000} bg}                                 & motorbike                                                    & \multirow{-2}{*}{-}                                    \\ \hline
Img3                                                     & car                                                             & -                                                         & -                                                            & car                                                    \\ \hline
                                                         & person                                                          & person                                                    &                                                              & {\color[HTML]{3531FF} bg}                              \\
\multirow{-2}{*}{Img4}                                   & car                                                             & {\color[HTML]{FE0000} bg}                                 & \multirow{-2}{*}{-}                                          & car                                                    \\ \hline
Img5                                                     & person                                                          & person                                                    & -                                                            & -                                                      \\ \hline
\end{tabular}}
    \end{subtable}%
    \begin{subtable}{.33\linewidth}
      \centering
        \caption{Disjoint}
        \resizebox{0.95\textwidth}{!}{\begin{tabular}{cc|ccc}
\hline
\cellcolor[HTML]{EFEFEF}                                 & \cellcolor[HTML]{EFEFEF}                                        & \multicolumn{3}{c}{\cellcolor[HTML]{EFEFEF}\textbf{Given ground truth class}}                                                                                                     \\
\multirow{-2}{*}{\cellcolor[HTML]{EFEFEF}\textbf{Image}} & \multirow{-2}{*}{\cellcolor[HTML]{EFEFEF}\textbf{Oracle class}} & \begin{tabular}[c]{@{}c@{}}Task 1\\ (person)\end{tabular} & \begin{tabular}[c]{@{}c@{}}Task 2\\ (motorbike)\end{tabular} & \begin{tabular}[c]{@{}c@{}}Task 3\\ (car)\end{tabular} \\ \hline
                                                         & person                                                          & {\color[HTML]{333333} }                                   & {\color[HTML]{333333} }                                      & {\color[HTML]{3531FF} bg}                              \\
                                                         & motorbike                                                       & {\color[HTML]{333333} }                                   & {\color[HTML]{333333} }                                      & {\color[HTML]{3531FF} bg}                              \\
\multirow{-3}{*}{Img1}                                   & car                                                             & \multirow{-3}{*}{{\color[HTML]{333333} -}}                & \multirow{-3}{*}{{\color[HTML]{333333} -}}                   & car                                                    \\ \hline
                                                         & person                                                          &                                                           & {\color[HTML]{3531FF} bg}                                    &                                                        \\
\multirow{-2}{*}{Img2}                                   & motorbike                                                       & \multirow{-2}{*}{-}                                       & motorbike                                                    & \multirow{-2}{*}{-}                                    \\ \hline
Img3                                                     & car                                                             & -                                                         & -                                                            & car                                                    \\ \hline
                                                         & person                                                          &                                                           &                                                              & {\color[HTML]{3531FF} bg}                              \\
\multirow{-2}{*}{Img4}                                   & car                                                             & \multirow{-2}{*}{-}                                       & \multirow{-2}{*}{-}                                          & car                                                    \\ \hline
Img5                                                     & person                                                          & person                                                    & -                                                            & -                                                      \\ \hline
\end{tabular}}
    \end{subtable} 
    \begin{subtable}{.33\linewidth}
      \centering
        \caption{Partitioned (Ours)}
        \resizebox{0.95\textwidth}{!}{\begin{tabular}{cc|ccc}
\hline
\cellcolor[HTML]{EFEFEF}                                 & \cellcolor[HTML]{EFEFEF}                                        & \multicolumn{3}{c}{\cellcolor[HTML]{EFEFEF}\textbf{Given ground truth class}}                                                                                                     \\
\multirow{-2}{*}{\cellcolor[HTML]{EFEFEF}\textbf{Image}} & \multirow{-2}{*}{\cellcolor[HTML]{EFEFEF}\textbf{Oracle class}} & \begin{tabular}[c]{@{}c@{}}Task 1\\ (person)\end{tabular} & \begin{tabular}[c]{@{}c@{}}Task 2\\ (motorbike)\end{tabular} & \begin{tabular}[c]{@{}c@{}}Task 3\\ (car)\end{tabular} \\ \hline
                                                         & person                                                          & {\color[HTML]{333333} }                                   & {\color[HTML]{3531FF} bg}                                    & {\color[HTML]{3531FF} }                                \\
                                                         & motorbike                                                       & {\color[HTML]{333333} }                                   & {\color[HTML]{333333} motorbike}                             & {\color[HTML]{3531FF} }                                \\
\multirow{-3}{*}{Img1}                                   & car                                                             & \multirow{-3}{*}{{\color[HTML]{333333} -}}                & {\color[HTML]{FE0000} bg}                                    & \multirow{-3}{*}{{\color[HTML]{3531FF} -}}             \\ \hline
                                                         & person                                                          &                                                           & {\color[HTML]{3531FF} bg}                                    &                                                        \\
\multirow{-2}{*}{Img2}                                   & motorbike                                                       & \multirow{-2}{*}{-}                                       & motorbike                                                    & \multirow{-2}{*}{-}                                    \\ \hline
Img3                                                     & car                                                             & -                                                         & -                                                            & car                                                    \\ \hline
                                                         & person                                                          & person                                                    &                                                              & {\color[HTML]{333333} }                                \\
\multirow{-2}{*}{Img4}                                   & car                                                             & {\color[HTML]{FE0000} bg}                                 & \multirow{-2}{*}{-}                                          & \multirow{-2}{*}{{\color[HTML]{333333} -}}             \\ \hline
Img5                                                     & person                                                          & person                                                    & -                                                            & -                                                      \\ \hline
\end{tabular}}
    \end{subtable} 
\end{table}

Similar to data configuration in CIL, the \textit{disjoint} scenario in CISS ensures the separation of data from each task. However, achieving disjointness in the \textit{disjoint} scenario requires prior knowledge of unseen classes and excludes data that could cause background shifts of these unseen classes (indicated by the absence of \textcolor{red}{bg} in the table). Given that background shift poses a significant challenge in CISS, many studies opt for the \textit{overlapped} scenario, which can result in background shifts of both previous and unseen classes. However, in this work, we highlight the issues associated with overlapping data in the \textit{overlapped} scenario and propose a novel scenario called \textit{partitioned}. Our proposed scenario ensures the separation between task data while accommodating background shifts of both unseen and previous classes.

\newpage

\subsection{Details for pseudo-labeling analysis}\label{sec: appendix pseudo analysis}
\subsubsection{Overview of Pseudo Retrieval Rate (PRR)}
\begin{figure}[h]
\begin{center}
\includegraphics[scale=0.4]{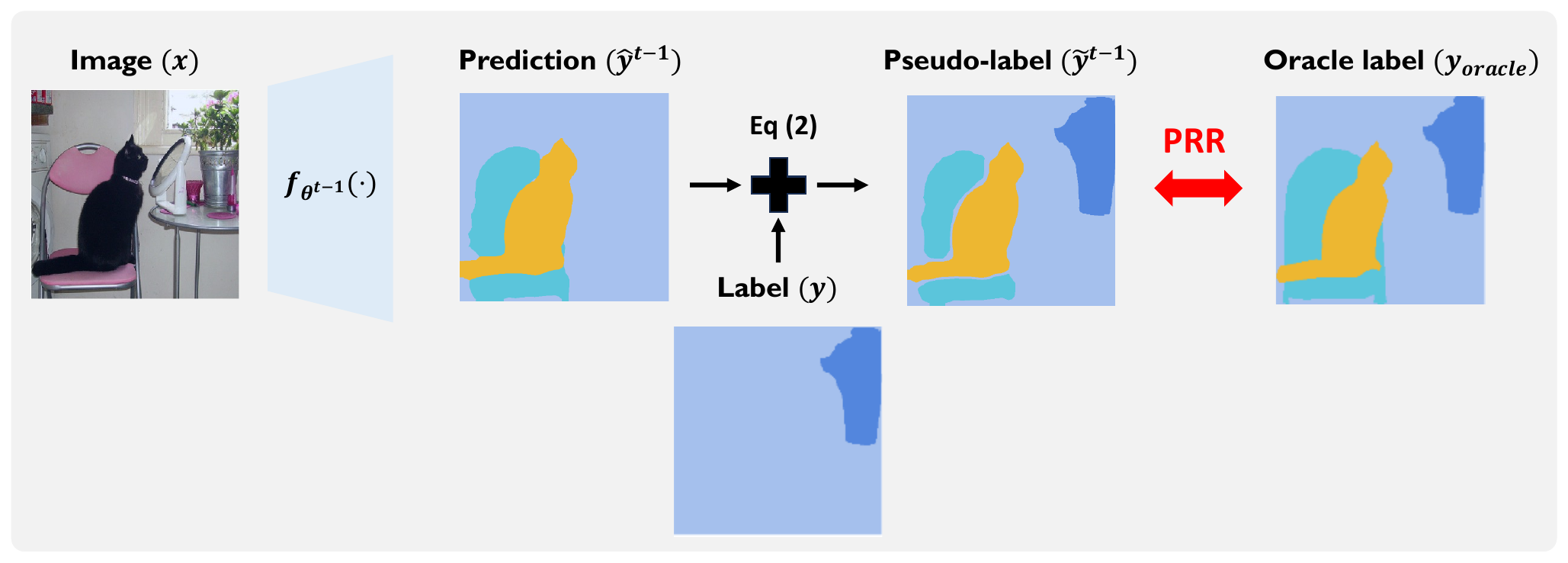}
\end{center}
\caption{An overview of Pseudo-labeling Retrieval Rate (PRR) metric evaluation}
\label{fig:appendix PRR overview details}
\end{figure}

Figure \ref{fig:appendix PRR overview details} illustrates the pseudo-labeling and the pseudo retrieval rate (PRR) evaluation procedure. First, the previously learned model $f_{\theta^{t-1}}$ predicts labels of the background region of a given image $x$. Then, the prediction $(\hat{y}^{t-1})$ is added on to the ground truth label $y$ to construct the pseudo-label $\tilde{y}^{t-1}$. After IoU for each class is calculated, the mean IoU over classes from the background class and the classes from the old task $\{c_{bg}\}\cup\mathcal{C}^{0:t-1}$ is returned.

\subsubsection{Implementation details for experiments}

\begin{table}[h]
\caption{Data configuration for pseudo-label analysis experiments in Section \ref{sec:rethinking scenario}}
\label{tab: appendix pseudo labeling analysis data}
\resizebox{0.5\textwidth}{!}{\begin{tabular}{cc|cccc}
    \toprule
     &             & \multicolumn{4}{c}{Number of each subset in $\mathcal{D}^{0}$}                                   \\ \cmidrule{3-6} 
     &             & \multirow{2}{*}{Non-overlapping} & \multicolumn{2}{c}{overlapping} & \multirow{2}{*}{Total} \\ \cmidrule{4-5}
Task & Class order &                                  & $\mathcal{D}_{seen}$        & $\mathcal{D}_{unseen}$        &                        \\ \midrule \midrule
15-1 & Type a      & 9234                             & 167           & 167             & 9568                   \\
     & Type b      & 9214                             & 32            & 31              & 9277                   \\
     & Type c      & 8449                             & 197           & 197             & 8843                   \\ \midrule
15-5 & Type a      & 8437                             & 566           & 565             & 9568                   \\
     & Type b      & 8452                             & 413           & 412             & 9277                   \\
     & Type c      & 7506                             & 669           & 668             & 8843      
    \tabularnewline \bottomrule
\end{tabular}}
\end{table}

\noindent{\textbf{Data configuration}} The base task dataset $\mathcal{D}^{0}$ consists of overlapping dataset reintroduced in future tasks and non-overlapping dataset which solely appear in task $0$. In the experiment, the overlapping dataset is randomly divided into half, $\mathcal{D}_{seen}$ and $\mathcal{D}_{seen}$. Table \ref{tab: appendix pseudo labeling analysis data} shows the number of each dataset, non-overlapping, $\mathcal{D}_{seen}$, $\mathcal{D}_{seen}$, and $\mathcal{D}^{0}$. The model is trained on the modified training dataset $\mathcal{D}^{0} \setminus \mathcal{D}_{unseen}$, and then PRR evaluation is done on $\mathcal{D}_{unseen}$ and $\mathcal{D}_{seen}$.

\noindent{\textbf{Training details}} The fine-tuning model is trained under the same training scheme used for PLOP in \citep{douillard2021plop}. For training details, please refer PLOP details in Table \ref{tab: appendix training details}. 

\subsection{Details for exemplar memory analysis}\label{sec: appendix memory analysis}

\subsubsection{Implementation details for experiment}

\noindent{\textbf{Data configuration}} After the model is trained with $\mathcal{D}^{0}$ at task $0$, the model is fine-tuned on data from $\mathcal{D}^{1}$ and the exemplar memory at task $1$. To construct an \textit{overlapping memory}, we first calculate the average overlapping ratio in the memory which indicates the total amount of data from $\mathcal{D}^{0}\cap\mathcal{D}^{1}$ out of the exemplar memory size $ M$. Through $100$ different class-balanced memory construction, we report the average ratio for each task and class order. Table \ref{tab: appendix memory analysis data} shows the number of training data used in each task and the ratio of overlapping data in the \textit{overlapping memory}.



\noindent{\textbf{Training details}} The fine-tuning model is trained under the same training scheme used for PLOP in \citep{douillard2021plop}. For training details, please refer PLOP details in Table \ref{tab: appendix training details}.

\begin{table}[h]
\caption{Data configuration for each dataset when \textit{overlapping memory} is used for exemplar memory analysis experiment in Section \ref{sec:rethinking scenario}}
\label{tab: appendix memory analysis data}
\resizebox{0.7\textwidth}{!}{
\begin{tabular}{ccc|cccc}
\toprule
Task & Class order & \begin{tabular}[c]{@{}c@{}}Overlapping ratio \\ in memory\end{tabular} & $\mathcal{D}^{0}$  & $\mathcal{D}^{0} \cap \mathcal{D}^{1}$ & $\mathcal{D}^{1}$  & \begin{tabular}[c]{@{}c@{}}$\mathcal{M}^{0}$\\ (Non-overlapping/overlapping)\end{tabular} \\ \midrule \midrule
15-1 & Type a      & 3.42 \%              & 9568 & 334                        & 487  & 97 / 3                                                                                  \\
     & Type b      & 3.48 \%              & 9214 & 394                        & 1177 & 97 / 3                                                                                  \\
     & Type c      & 16.6 \%             & 8672 & 2425                       & 3898 & 83 / 17                                                                                 \\ \midrule
15-5 & Type a      & 9.58 \%             & 9568 & 1131                       & 2145 & 90 / 10                                                                                 \\
     & Type b      & 8.81 \%              & 8843 & 1337                       & 3076 & 91 / 9                                                                                  \\
     & Type c      & 35.3 \%             & 8672 & 3045                       & 4955 & 65 / 35                                                                                 \\ \midrule
10-1 & Type a      & 5.61 \%              & 6139 & 431                        & 528  & 94 / 6                                                                                  \\
     & Type b      & 0.36 \%              & 7703 & 48                         & 264  & 99 / 1                                                                                  \\
     & Type c      & 35.81 \%             & 5255 & 1831                       & 3898 & 64 / 36                                                                                 \\ \midrule
10-5 & Type a      & 34.22 \%             & 6139 & 2113                       & 5542 & 66 / 34                                                                                 \\
     & Type b      & 16.23 \%             & 7703 & 1523                       & 2663 & 84 / 16                                                                                 \\
     & Type c      & 45.58 \%             & 5255 & 2353                       & 6406 & 54 / 46                                                                             
     \tabularnewline \bottomrule
\end{tabular}}
\end{table}

\newpage

\subsection{Implementation error in previous stuides}

\begin{figure}[h]
\begin{center}
\includegraphics[scale=0.4]{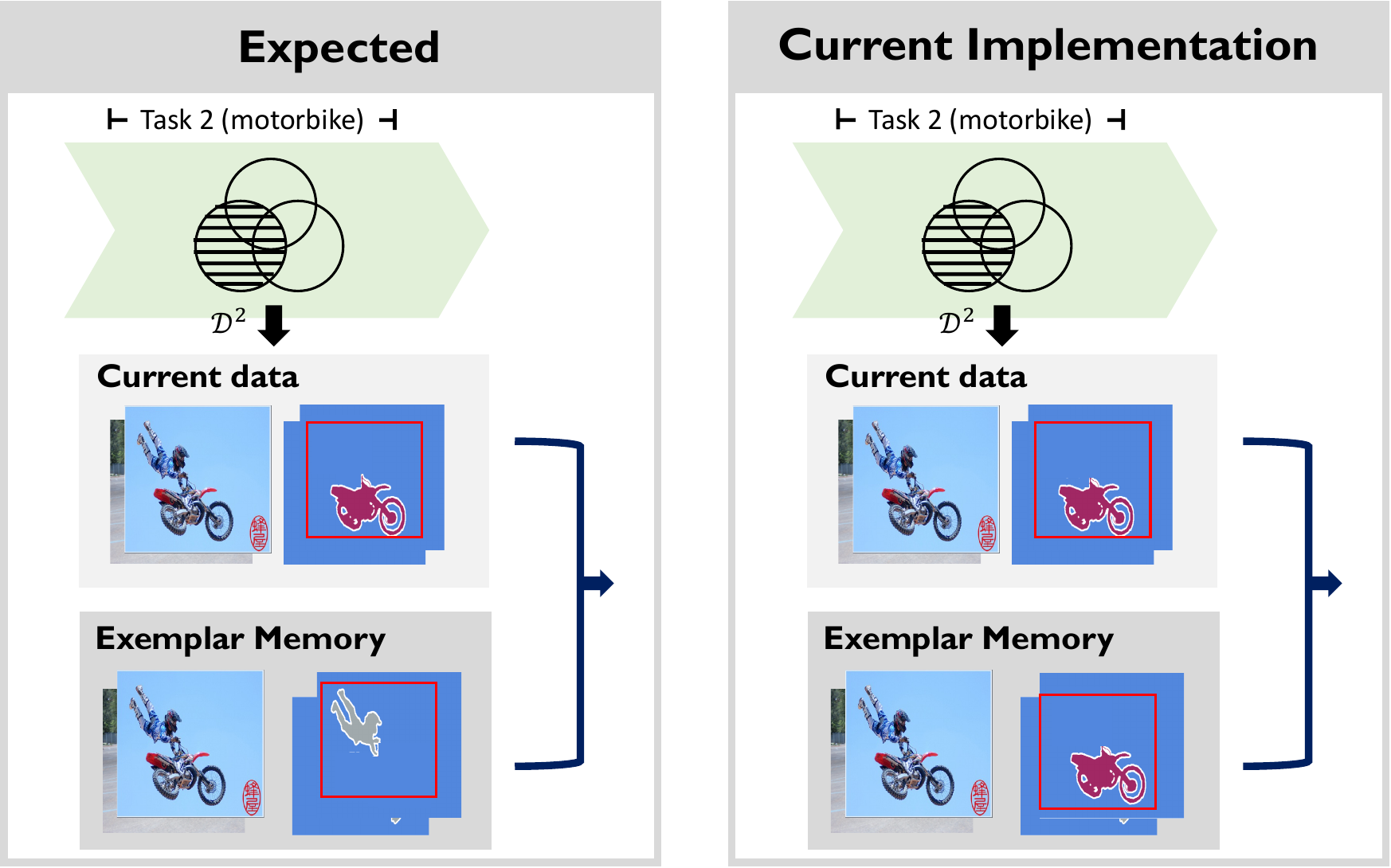}
\end{center}
\caption{This figure illustrates the labeling issue in code implementation of current CISS studies \citep{cha2021ssul, baek2022decomposed, zhang2022mining, zhang2023coinseg}. Note that the ground-truth mask of data from the exemplar memory does not provide labels for the classes of the previous task at which it was stored. Instead, it provides labels for the classes of the current task.}
\label{fig:appendix target labeling for memory error}
\end{figure}

As illustrated in Figure \ref{fig:appendix target labeling for memory error}, the model is expected to see ground-truth masks annotated with \textit{motorbike} class for current data and \textit{person} class for data stored in memory, respectively. However, following code implementation of previous studies \citep{cha2021ssul, baek2022decomposed, zhang2022mining, zhang2023coinseg}, the ground-truth mask for memory data is labeled with \textit{motorbike} class.

\newpage

\subsection{Implementation of MiB-AugM, PLOP-M, and $\text{DKD-M}^{\dagger}$}\label{sec: appendix MiB-M, DKD-M, PLOP-M}
In this section, we explain the concrete formula of MiB-AugM, PLOP-M, and $\text{DKD-M}^{\dagger}$ following the notation in Section \ref{sec:notation}.

\subsubsection{MiB-AugM}

The overall loss for MiB-AugM model can be defined as follows.

\begin{equation}
    \label{eq:MiB-AugM method}
    \begin{aligned}
     \mathcal{L}(\theta^{t})  =\underbrace{\frac{1}{|\mathcal{D}^{t}|}\sum_{(x,y)\in \mathcal{D}^{t}}\mathcal{L}_{unce}(y, f_{\theta^{t}}(x)) +\frac{\lambda}{|\mathcal{D}^{t}\cup\mathcal{M}^{t-1}|}\sum_{(x,y)\in \mathcal{D}^{t}\cup\mathcal{M}^{t-1}}\mathcal{L}_{unkd}(f_{\theta^{t-1}}(x), f_{\theta^{t}}(x))}_\text{From MiB \citep{cermelli2020modeling}} \\  + \underbrace{\frac{1}{|\mathcal{M}^{t-1}|}\sum_{(x,y)\in\mathcal{M}^{t-1}}\mathcal{L}_{mem}(y, f_{\theta^{t}}(x))}_\text{Our proposed memory loss}
    \end{aligned}
\end{equation}

Here, we rewrite the formula of $\mathcal{L}_{unce}$ and $\mathcal{L}_{unkd}$ defined in \citet{cermelli2020modeling} on our notation.

\begin{equation}
    \label{eq:unce}
    \begin{aligned}
    \mathcal{L}_{unce}(y, f_{\theta^{t}}(x)) = -\frac{1}{N}\sum_{i=1}^{N}\log \ddot{p}^{t}_{i,y_{i}}
    \end{aligned}
\end{equation}

\begin{equation}
    \begin{aligned}
    \mathcal{L}_{unkd}(f_{\theta^{t-1}}(x), f_{\theta^{t}}(x)) = -\frac{1}{N}\sum_{i=1}^{N}{\sum_{c\in \mathcal{C}^{0:t-1}}{{p}^{t-1}_{i,c}\log \dot{p}^{t}_{i,c}}}
    \end{aligned}
\end{equation}

where $\ddot{p}^{t}_{i,c}$ and $\dot{p}^{t}_{i,c}$ indicates different augmentation technique of prediction probabilities. The augmented prediction for $\mathcal{L}_{unce}$, $\ddot{p}^{t}_{i,c}$, is defined as follows.

\begin{equation}
    \ddot{p}_{i,c}^{t}=\begin{cases}
    p_{i,c}^{t}  & \quad \text{if } c \neq c_{bg}
    \\
    p_{i,c_{bg}}^{t}+\sum_{k\in\mathcal{C}^{0:t-1}}p_{i,k}^{t}
      & \quad    \text{if }  c=c_{bg}
    \end{cases}
    \label{eq:mem prediction}
\end{equation}

\subsubsection{PLOP-M}
The loss function in PLOP-M remains unchanged, same as PLOP \citep{douillard2021plop}. However, PLOP-M distinguishes itself by updating with concatenated data from both the current task and memory, using an equal ratio from each.

\begin{equation}
    \label{eq:appendix plop-m}
    \begin{aligned}
    \mathcal{L}(\theta^{t})=\frac{1}{|\mathcal{D}^{t}\cup\mathcal{M}^{t-1}|}\sum_{(x,y)\in \mathcal{D}^{t}\cup\mathcal{M}^{t-1}}\mathcal{L}_{ce}(\tilde{y}, f_{\theta^{t}}(x))+\lambda\mathcal{L}_{pod}(f_{\theta^{t-1}}(x), f_{\theta^{t}}(x)) 
    \end{aligned}
\end{equation}
\subsubsection{$\text{DKD-M}^{\dagger}$}
The loss function in DKD-M in the original paper remains also unchanged in its original paper \citep{baek2022decomposed}. Namely, the data from concatenated set, \textit{i.e.,} $(x,y) \sim \mathcal{D}^{t}\cup\mathcal{M}^{t-1}$, was forwarded to $\mathcal{L}_{kd}. \mathcal{L}_{dkd}, \mathcal{L}_{mbce}$, and $\mathcal{L}_{ac}$. However, looking at $\mathcal{L}_{mbce}$ in eq \ref{eq:appendix mbce}, it only updates the new class score, which is awkward for data in memory that does not have any labels of new classes. 

\begin{equation}
\label{eq:appendix mbce}
    \begin{aligned}
    \mathcal{L}_{mbce}(y, f_{\theta^{t}}(x)) = -\frac{1}{N}\sum_{i=1}^{N}{\sum_{c\in \mathcal{C}^{t}}\gamma\1_{\{y_{i}=c\}}\log p^{t}_{i,c}+\1_{\{y_{i}\neq c\}}\log (1-p^{t}_{i,c})}
    \end{aligned}
\end{equation}

Later, we noticed that this wrong usage did not harm the performance because the labeling issue existed in memory retrieval mentioned in Section \ref{sec:implementation} (Manuscript). Therefore, after correction in memory target labeling, we add a $\mathcal{L}_{mbce}$ loss for memory, dubbed as $\mathcal{L}_{membce}$, which is defined as follows.

\begin{equation}
\label{eq:appendix membce}
    \begin{aligned}
    \mathcal{L}_{membce}(y, f_{\theta^{t}}(x)) = -\frac{1}{N}\sum_{i=1}^{N}{\sum_{c\in \mathcal{C}^{0:t-1}}\gamma\1_{\{y_{i}=c\}}\log p^{t}_{i,c}+\1_{\{y_{i}\neq c\}}\log (1-p^{t}_{i,c})}
    \end{aligned}
\end{equation}

The overall loss function for $\text{DKD-M}^{\dagger}$ is as follows.

\begin{equation}
    \label{eq:appendix dkd-m}
    \begin{aligned}
    \mathcal{L}(\theta^{t})=\frac{1}{|\mathcal{D}^{t}\cup\mathcal{M}^{t-1}|}\sum_{(x,y)\in \mathcal{D}^{t}\cup\mathcal{M}^{t-1}}\underbrace{\bigg[\alpha\mathcal{L}_{kd}(f_{\theta^{t-1}}(x), f_{\theta^{t}}(x))+\beta\mathcal{L}_{dkd}(f_{\theta^{t-1}}(x), f_{\theta^{t}}(x))\bigg]}_\text{From DKD \citep{baek2022decomposed}} \\+\frac{1}{|\mathcal{D}^{t}|}\sum_{(x,y)\in \mathcal{D}^{t}}{\underbrace{\bigg[\mathcal{L}_{mbce}(y, f_{\theta^{t}}(x))+\mathcal{L}_{ac}(y, f_{\theta^{t}}(x))\bigg]}_\text{From DKD \citep{baek2022decomposed}}} \\ + \frac{1}{|\mathcal{M}^{t-1}|}\sum_{(x,y)\in\mathcal{M}^{t-1}}\underbrace{\mathcal{L}_{membce}(y, f_{\theta^{t}}(x))}_\text{Added loss modified from $\mathcal{L}_{mbce}$} 
    \end{aligned}
\end{equation}

\newpage

\subsection{Implementation details}\label{sec: appendix hyperparameters}

\begin{table}[h]
\caption{The number of train data for every task in the \textit{overlapped} and \textit{partitioned} scenarios.}
\centering
\scalebox{0.7}{\begin{tabular}{c|c|c|c|c}
\toprule \toprule
\textbf{Task}                       & \textbf{Scenario}                     & \textbf{Seed}   & \textbf{The number of training data for each task}                                  & \textbf{Total} \\ \midrule
\multirow{4}{*}{15-1 Task} & overlapped                   & -      & 9568 / 487 / 299 / 491 / 500 / 548                                 & 11893 \\
                           & \multirow{3}{*}{partitioned} &  0 & 9031 / 254 / 266 / 270 / 434 / 327                                 & 10582 \\
                           &                              &  1 & 9059 / 263 / 271 / 256 / 413 / 320                                 & 10528 \\
                           &                              &  2 & 9030 / 274 / 258 / 253 / 437 / 330                                 & 10582 \\ \midrule
\multirow{4}{*}{5-3 Task}  & overlapped                   & -      & 2836 / 2331 / 1542 / 2095 / 4484 / 1468                            & 14756 \\
                           & \multirow{3}{*}{partitioned} &  0 & 2222 / 1860 / 972 / 1574 / 2923 / 1031                             & 10582 \\
                           &                              &  1 & 2237 / 1859 / 991 / 1616 / 2890 / 989                              & 10582 \\
                           &                              &  2 & 2221 / 1855 / 993 / 1589 / 2904 / 1020                             & 10582 \\ \midrule
\multirow{4}{*}{10-1 Task} & overlapped                   & -      & 6139 / 528 / 1177 / 444 / 482 / 3898 / 487 / 299 / 491 / 500 / 548 & 14993 \\
                           & \multirow{3}{*}{partitioned}                  &  0 & 4847 / 207 / 961 / 310 / 303 / 2403 / 254 / 266 / 270 / 434 / 327  & 10582 \\ &                              &  1 & 4861 / 226 / 987 / 322 / 307 / 2356 / 263 / 271 / 256 / 413 / 320                              & 10582 \\
                           &                              &  2 & 4856 / 213 / 969 / 317 / 303 / 2372 / 274 / 258 / 253 / 437 / 330                             & 10582 \\ \midrule
\multirow{4}{*}{10-5 Task} & overlapped                   & -      & 6139 / 5542 / 2145                                                 & 13826 \\
                           & \multirow{3}{*}{partitioned}                  &  0 & 4847 / 4148 / 1551                                                 & 10582 \\ &                              &  1 & 4861 / 4198 / 1523                              & 10582 \\
                           &                              &  2 & 4856 / 4174 / 1552                             & 10582 \\ 
                           \bottomrule \bottomrule
\end{tabular}}
\label{tab:appendix training data numbers}
\end{table}

\noindent{\textbf{Baseline reproduction and experimental environment}} The code environment of CISS is divided into two branches: Distributed data parallel (DDP) implemented on Nvidia Apex (\url{https://github.com/NVIDIA/apex}) and Torch \citep{paszke2017automatic}. Initial studies primarily utilized the former, with \citet{csseg2023} organizing numerous baselines for evaluation. Due to the transition of Nvidia Apex to Torch in deep learning community\footnote{Please refer to \url{https://github.com/NVIDIA/apex/issues/818}}, recent CISS works began to work on Torch environment. However, recent works conducted on the latter omitted the process of re-implementing baselines, instead reporting figures from the original paper. Given the transition from Apex to Torch, resulting in significant changes in built-in operations, it's widely acknowledged that a mismatch in results exists between the two environments. To facilitate fair comparisons, we report our implementation of two baselines in the Torch version.

\noindent{\textbf{Dataset and protocols}} Pascal VOC 2012 \citep{everingham2010pascal} consists of 10.582 training and 1,449 validation images for 20 object and background classes. For the training image, random crop with size $512$, random resize with $(0.5, 2.0)$ ratio, and normalization are used. For the test image, only normalization is used.

Following \citet{cermelli2020modeling}, incremental tasks are denoted by the number of classes used in base classes, $|\mathcal{C}^{0}|$, and the number of classes learned in each incremental task, $|\mathcal{C}^{t}| \text{ }\forall t \in \{1,\dots, T\}$. For example, if the base task class is composed of $15$ classes and $1$ class is learned at every task, it is denoted as $15\text{-}1$ task.

Table \ref{tab:appendix training data numbers} demonstrates the number of data $\mathcal{D}^{t}$ used for each task in both the \textit{partitioned} and \textit{overlapped} scenarios. Note that the dataset remains consistent across tasks in the \textit{overlapped}, whereas variations are observed in the \textit{partitioned} scenario across different seeds.

\noindent{\textbf{Training details}} Table \ref{tab: appendix training details} summarizes the training details used for each method. MiB, PLOP-M, and DKD-M use the same details as MiB-AugM, PLOP, and DKD, respectively. Note that some training details of MiB are different from those that were used in MiB. For methods that use exemplar memory, we replace half of the data from the current batch with the data from the memory.

\begin{table}[h]
\caption{Training details for each method}
\label{tab: appendix training details}
\resizebox{\textwidth}{!}{\begin{tabular}{l|ccccc|cc|cc}
\toprule \toprule
         & \multicolumn{5}{c|}{\textbf{Common}}                                                                      & \multicolumn{2}{c|}{\textbf{Base task}}                                                    & \multicolumn{2}{c}{\textbf{Incremental task}}                                                                                                                                \\
         & Batch size & Epoch & Optimizer & \multicolumn{1}{l}{Momentum} & \multicolumn{1}{l|}{Lr schedule} & \begin{tabular}[c]{@{}c@{}}Lr\\ (backbone / aspp / classifier)\end{tabular} & Weight decay & \begin{tabular}[c]{@{}c@{}}Lr\\ (backbone / aspp / classifier)\end{tabular} & \begin{tabular}[c]{@{}c@{}}Weight decay\\ (backbone / aspp / classifier)\end{tabular} \\ \midrule
PLOP \scriptsize \citep{douillard2021plop}    & 24         & 30    & SGD       & Nesterov, 0.9                & PolyLR                           & 0.01 / 0.01 / 0.01                                                          & 0.001        & 0.001 / 0.001 / 0.001                                                       & 0.001 / 0.001 / 0.001                                                                 \\
DKD \scriptsize \citep{baek2022decomposed}     & 32         & 60    & SGD       & Nesterov, 0.9                & PolyLR                           & 0.001 / 0.01 / 0.01                                                         & 0.0001       & 0.0001 / 0.001 / 0.001                                                      & 0 / 0 / 0.0001                                                                        \\
MiB-AugM (Ours) & 24         & 30    & SGD       & Nesterov, 0.9                & PolyLR                           & 0.01 / 0.01 / 0.01                                                          & 0.001        & 0.001 / 0.01 / 0.01                                                         & 0 / 0 / 0.001                                                                         \\ \bottomrule \bottomrule
\end{tabular}}

\end{table}

\subsubsection{hyper-parameters}

Table \ref{tab: appendix hyperparam table} summarizes the main hyper-parameters of each method used for training. Notations for each hyper-parameter are from the original paper. We also used same hyper-parameter used in MiB \citep{cermelli2020modeling}.

\begin{table}[h]
\caption{Hyper-parameters used for each method}
\label{tab: appendix hyperparam table}
\resizebox{\textwidth}{!}{\begin{tabular}{l|c|c|c}
\toprule \toprule
\multicolumn{1}{c|}{} & \textbf{Base task training} & \textbf{Incremental task} & \multicolumn{1}{c}{\textbf{Inference}} \\ \midrule
PLOP \scriptsize \citep{douillard2021plop}                 &          -                   &  $\lambda_{f}=0.01$ (features), $\lambda_{l}=0.0005$ (logits), $\tau=0.001$, pod scale$=[1,\frac{1}{2},\frac{1}{4}]$                         &             -                           \\
DKD  \scriptsize \citep{baek2022decomposed}                 &        $\gamma=2$                     &         $\alpha=5, \beta=5, \gamma=1$                   &      $\tau=0.5$                                  \\
MiB-AugM (Ours)             &   -                          &        $\lambda=5$                   &        -                                \\ \bottomrule \bottomrule
\end{tabular}}

\end{table}

\newpage

\subsection{Computation details}

The experiments were conducted using PyTorch \citep{paszke2017automatic} 1.13.1 with CUDA 11.2 and were run on four NVIDIA Titan XP GPUs with 12GB memory per device. All experiments except PLOP \citep{douillard2021plop} were conducted with distributed data-parallel training on four GPUs. For PLOP \citep{douillard2021plop} experiments, we use 2 GPUs for parallel training since the results of the original paper could not be achieved by other numbers.

\subsection{Software and Dataset Licenses}

\subsubsection{Datasets}
\begin{itemize}
    \item{Pascal VOC \citep{everingham2010pascal}: CC BY-NC-SA 3.0 License} \\ \url{http://host.robots.ox.ac.uk/pascal/VOC/}
\end{itemize}
\subsubsection{Models}
\begin{itemize}
    \item{MiB \citep{cermelli2020modeling}: MiT License} \\ \url{https://github.com/fcdl94/MiB}
    \item{PLOP \citep{douillard2021plop}: MiT License } \\ \url{https://github.com/arthurdouillard/CVPR2021_PLOP}
    \item{DKD \citep{baek2022decomposed}: GPL-3.0 License} \\ \url{https://github.com/cvlab-yonsei/DKD}
\end{itemize}

\end{document}